\documentclass[11pt]{article}
\usepackage[T1]{fontenc}
\usepackage[utf8]{inputenc}
\usepackage{textcomp}
\usepackage{amsmath,amssymb}
\usepackage{graphicx}
\usepackage{longtable,booktabs,array,calc}
\usepackage{url}
\usepackage[hidelinks]{hyperref}
\usepackage[margin=1in]{geometry}
\usepackage{microtype}
\usepackage{enumitem}
\setlist{nosep}
\hypersetup{pdfauthor={Snigdhajyoti Rakshit},pdftitle={Deep Learning for Satellite-Based Mountain Hazard Screening: A Retrospective Failure Analysis on the 2021 Chamoli Rock-Ice Avalanche}}
\DeclareUnicodeCharacter{00D7}{\ensuremath{\times}}
\DeclareUnicodeCharacter{00B3}{\textsuperscript{3}}
\DeclareUnicodeCharacter{00B2}{\textsuperscript{2}}
\DeclareUnicodeCharacter{00B9}{\textsuperscript{1}}
\DeclareUnicodeCharacter{2070}{\textsuperscript{0}}
\DeclareUnicodeCharacter{2074}{\textsuperscript{4}}
\DeclareUnicodeCharacter{2076}{\textsuperscript{6}}
\DeclareUnicodeCharacter{2077}{\textsuperscript{7}}
\DeclareUnicodeCharacter{207B}{\textsuperscript{-}}
\DeclareUnicodeCharacter{207B}{\textsuperscript{-}}
\DeclareUnicodeCharacter{2212}{\ensuremath{-}}
\DeclareUnicodeCharacter{00B0}{\ensuremath{^\circ}}
\DeclareUnicodeCharacter{03C1}{\ensuremath{\rho}}
\DeclareUnicodeCharacter{0394}{\ensuremath{\Delta}}
\DeclareUnicodeCharacter{03A3}{\ensuremath{\Sigma}}
\DeclareUnicodeCharacter{03B2}{\ensuremath{\beta}}
\DeclareUnicodeCharacter{03B8}{\ensuremath{\theta}}
\DeclareUnicodeCharacter{03C6}{\ensuremath{\phi}}
\DeclareUnicodeCharacter{03B1}{\ensuremath{\alpha}}
\DeclareUnicodeCharacter{2265}{\ensuremath{\geq}}
\DeclareUnicodeCharacter{2264}{\ensuremath{\leq}}
\DeclareUnicodeCharacter{2260}{\ensuremath{\neq}}
\DeclareUnicodeCharacter{221A}{\ensuremath{\sqrt{}}}
\DeclareUnicodeCharacter{03C3}{\ensuremath{\sigma}}
\DeclareUnicodeCharacter{03C0}{\ensuremath{\pi}}
\DeclareUnicodeCharacter{00BD}{\ensuremath{\frac{1}{2}}}
\DeclareUnicodeCharacter{00B1}{\ensuremath{\pm}}
\DeclareUnicodeCharacter{00A9}{\textcopyright}
\begin{document}
\title{Deep Learning for Satellite-Based Mountain Hazard Screening: A Retrospective Failure Analysis on the 2021 Chamoli Rock-Ice Avalanche}
\author{Snigdhajyoti Rakshit\\
\small School of Computing Science and Engineering, VIT Bhopal University, India\\
\small \href{mailto:snigdhajyotirakshit@gmail.com}{snigdhajyotirakshit@gmail.com};
\href{mailto:snigdhajyoti.23bce11609@vitbhopal.ac.in}{snigdhajyoti.23bce11609@vitbhopal.ac.in}\\
\small ORCID: \href{https://orcid.org/0009-0006-8132-5668}{0009-0006-8132-5668}}
\date{}
\maketitle
\begin{abstract}

The approach utilizes 10 bands from Sentinel-2 imagery, along with terrain information, in a deep learning pipeline to classify potential hazard sources in any given mountainous region. The performance of our model has been evaluated against the data collected during the rock-ice avalanche event in the Chamoli region in the Indian Himalayas on 7th February 2021. U-Net performs segmentation of the detected land surface in the satellite image and detects changes to identify candidate hazard zones, and then the following steps are performed to calculate the potential energy using the Monte Carlo sampling for steepest descent (D8) flow paths, finding the exposed structures in the downstream locations, and ranking the zones depending on the risk score. In comparison to the simple spectral rule, our approach achieved the higher macro F1 score of 0.648 (as compared to 0.519). But the F1 score in case of rock and soil was low, being in the range of 0.3s. In case of optical change detection in winter, it's only that northeast-facing slope which couldn't be detected, as it was under 73\% of shade prior to the incident. The parameters have been set for this particular scene alone and not independently; hence, this is a retrospective test, and this test locates a failure which has occurred earlier. The primary contribution of this paper is locating the pipeline leak.

\end{abstract}


\section{Introduction}\label{introduction}

Most mountain disasters occur high up on the mountain before descending into the valleys as a result of the falling of rocks, hanging glacier breaking off, moraine lake bursting or flow of a mass of debris in channels resembling that of a fluid. On 7 February 2021, roughly 27 × 106 m³ of rocks and Approximately 27 × 106 m³ of rocks and glacier ice collapsed from Ronti Peak and developed into a debris flow. This disaster has not only devastated the Rishiganga Project but has also caused damage to the Tapovan Vishnugad Hydroelectric Project, resulting in the loss of at least 200 lives {[}3, 4{]}.

Images taken by the satellite show similar terrains repeatedly {[}5{]}. But changes in the terrains are not an appropriate way to determine the amount of material movement and its destination. Further, it is not clear whether there are people who live in the affected region under the slope. All the deep learning-based landslide detection models aim to generate a change map of the affected regions. In this study, we are evaluating the screening pipeline based on which the information from various sources will be converted into physical properties for the four different categories of evidence. This solution designed to aid in identifying areas that may need further analysis by a specialist. This pipeline was meant to prioritize sites for specialist analysis and not for failure prediction. The test case is a retrospective one, as it uses data collected after the event (DEM of February 2021) to identify a source which has already failed and thus cannot provide advance warning of a failure.

For this case study, the pipeline analysis is done on an actual event, and four key questions are asked:

\begin{enumerate}
\def\labelenumi{\arabic{enumi}.}
\item
  Is it possible that one of the segmentation algorithms generates more appropriate surfaces for change detection compared to the spectral thresholds in a different geographic location?
\item
  Does the routing algorithm generate an equivalent routing from the DEM using the described source that matches the actual flow path and exposure?
\item
  Are there any sources found? If the source is identified, which zone does the pipeline consider the best near the source, and how have all the elements contributed to the risk assessment?
\item
  What fails, and why?
\end{enumerate}

First, we develop a minimal screening pipeline that we describe in terms of its procedure. Next, we assess retrospectively the screening results of the Chamoli 2021 event based on a documented reference event with a consistent set of evaluation rules. Nevertheless, the configuration of the model was not independent from the earlier screenings in the same scene (Section 5.8).

Furthermore, we developed some additional analysis steps: 1) Difference of the published 2 m DEM for volume, 2) Ranking and component removal, 3) Change accuracy of a stratified point sample, and 4) ERA5 climate indicators. We will discuss in detail how and why our screening pipeline fails at the moment. Almost all of these shortcomings result from a compromise between two goals: eliminating shadow change and detecting extremely cold north-facing areas in the winter image.

\section{Related Work}\label{related-work}

The collapse was analyzed by Shugar et al. {[}3{]} through satellite imagery, seismograms, and pre- and post-collapse DEMs. The detachment took place near the elevation of 5,500 m on the north-facing cliff of Ronti Peak (the slope at 1 km distance from the source within the present DEM faces in a northeast direction; see Section 5.4). The estimated volume is 26.9 × 10⁶ m³ (95\% confidence interval of 26.5 to 27.3 × 10⁶ m³), and the flow traveled 15 km to the Rishiganga project and 26 km to Tapovan from the source. The 2 m resolution September 2015 and February 2021 DEM composites by Shean et al. {[}8{]} were used. The GEER reconnaissance report {[}9{]} specified the locations and coordinates of the affected areas.

The U-Net {[}10{]} is one of the popular architectures for the dense labeling of an image using encoder-decoder framework. It is extensively used for the Sentinel-2 land cover classification, with or without sparse labels {[}11{]} or multi-date images {[}12{]}. We compare class maps for two dates after individual classification of them. Post-classification comparison is easy to implement and interpret but it captures the errors of both dates {[}13{]}. Spectral index and change vector differencing are typical baselines for the change detection {[}14{]}.

While the D8 algorithm {[}15{]} is often employed in defining a unique downslope flow for a particular pixel, regional susceptibility approaches such as Flow-R {[}16{]} allocate flow to multiple downslope pixels. For estimating the error of DEM differences volumes, it must be calculated from robust statistics from stable regions; one example is normalized median absolute deviation {[}17{]}. Stratified area-weighted sampling {[}18{]} is advised for determining accuracy and extent of land change maps, and that's how we proceed with change detection.

The current state of research for satellite-based landslide detection uses deep learning-based segmentation. Landslide4Sense, an open benchmark dataset, relies on Sentinel-2 imagery together with DEM and slope. SAR-based detectors also have the capability of working under conditions of cloudiness and low light levels. The current state of art approaches mainly relies on the detection of ground changes. Glacial Lake Susceptibility analysis relying on lake inventories and multi-criteria index approach, but they can deal with only one data source at a time. As mentioned, the literature search performed by us was not exhaustive, and there were no tests that used one pipeline from change detection, energy, routing, and exposure to ranked documented sources in component removal tests. Here, we present a test case for a single event. The most important result of our analysis is the proof of where the optical approach fails for this event.

\section{Study Area and Data}\label{study-area-and-data}

The scene is situated between 79.58° and 79.80° East, and 30.33° and 30.54° North (approximately 21 x 23 km) on 10 m grids in UTM zone 44N (EPSG:32644). It should be noted that these coordinates are approximate and don\textquotesingle t correspond to the source polygon described in the published source {[}3{]}. Only objects found within these coordinates that are included in the documented source and downstream checkpoints listed below can be viewed by the pipeline (Section 7). The total size of the scene after masking for both pre and post event imagery is 317.8 km². It constitutes around 65\% of the area of the bounding box or approximately 492 km². Data sources used for the current research and their purposes are presented in Table 1.

Table 1: Data sources and their use.

\begin{longtable}[]{@{}
  >{\raggedright\arraybackslash}p{(\columnwidth - 4\tabcolsep) * \real{0.3828}}
  >{\raggedright\arraybackslash}p{(\columnwidth - 4\tabcolsep) * \real{0.3467}}
  >{\raggedright\arraybackslash}p{(\columnwidth - 4\tabcolsep) * \real{0.2705}}@{}}
\toprule\noalign{}
\begin{minipage}[b]{\linewidth}\raggedright
\textbf{Data}
\end{minipage} & \begin{minipage}[b]{\linewidth}\raggedright
\textbf{Source}
\end{minipage} & \begin{minipage}[b]{\linewidth}\raggedright
\textbf{Use}
\end{minipage} \\
\midrule\noalign{}
\endhead
\bottomrule\noalign{}
\endlastfoot
Sentinel-2 L2A, t1 (January 16, 2021) {[}5, 20{]} & Microsoft Planetary Computer & pre-event surface map \\
Sentinel-2 L2A, t2 (February 25, 2021) & Planetary Computer & post-event surface map \\
Sentinel-2 L2A, t3 (March 12, 2022) & Planetary Computer & persistence (same season) \\
Copernicus GLO-30 DEM {[}21{]} & Planetary Computer; AWS open data (same product) & slope, illumination, routing \\
2 m DEM difference, September 2015 to February 10 and 11, 2021 & Shean et al. {[}8{]}, CC BY-NC 4.0 & follow-up source detection, volume \\
ERA5 daily reanalysis, 1991 to 2021 {[}22{]} & Open-Meteo archive API {[}23{]} & rainfall and temperature indicators \\
Settlements, bridges, hydropower & OpenStreetMap (45 features; 0 hydropower) {[}24{]} & exposure inventory \\
Reference event: source, 4 checkpoints, 3 damaged sites & GEER report {[}9{]}; Shugar et al. {[}3{]} & evaluation \\
\end{longtable}

The bands which we applied include B02, B03, B04, B08, B11 and B12. Both B11 and B12 are native 20 m bands and were resampled to the 10 m resolution. The cloud free day in the declared intervals for each date is chosen as follows: t1: from January 15 to February 6, 2021; t2: from February 8 to March 10, 2021; t3: from January 15 to March 15, 2022. The images with cloud cover above 30\% have been excluded from the analysis. t1 and t2 belong to the same winter period. As noted before, t1 image taken in February 2020 has been used to calculate the inter-annual difference in snow coverages. Sentinel-2 bands used: B02, B03, B04, B08, B11 and B12.

Labelling of polygons has been done through QGIS {[}25{]}, using true-color and SWIR false-color (B12-B08-B04) images of the acquisitions. Acquisition date(s), either t1, t2 or both have been noted for each polygon. Surface type of the polygons has been labelled as background, water, snow/ice, rock, soil or debris. In order to minimise the bias in selecting the test pixels for the test strip, instead of manually choosing the pixels, we have generated 30 random points on the bare ground, as well as 20 points on the steep bare ground. Each of the points was visually verified before being included in the data set. The final data set contains 63 polygons.

Table 2: Labeled pixels per class in the training, validation and test strips (t1 and t2 combined).

\begin{longtable}[]{@{}
  >{\raggedright\arraybackslash}p{(\columnwidth - 12\tabcolsep) * \real{0.2766}}
  >{\raggedright\arraybackslash}p{(\columnwidth - 12\tabcolsep) * \real{0.1633}}
  >{\raggedright\arraybackslash}p{(\columnwidth - 12\tabcolsep) * \real{0.0986}}
  >{\raggedright\arraybackslash}p{(\columnwidth - 12\tabcolsep) * \real{0.1449}}
  >{\raggedright\arraybackslash}p{(\columnwidth - 12\tabcolsep) * \real{0.1071}}
  >{\raggedright\arraybackslash}p{(\columnwidth - 12\tabcolsep) * \real{0.1071}}
  >{\raggedright\arraybackslash}p{(\columnwidth - 12\tabcolsep) * \real{0.1025}}@{}}
\toprule\noalign{}
\begin{minipage}[b]{\linewidth}\raggedright
\textbf{Strip (t1 + t2 pixels)}
\end{minipage} & \begin{minipage}[b]{\linewidth}\raggedright
\textbf{background}
\end{minipage} & \begin{minipage}[b]{\linewidth}\raggedright
\textbf{water}
\end{minipage} & \begin{minipage}[b]{\linewidth}\raggedright
\textbf{snow/ice}
\end{minipage} & \begin{minipage}[b]{\linewidth}\raggedright
\textbf{rock}
\end{minipage} & \begin{minipage}[b]{\linewidth}\raggedright
\textbf{soil}
\end{minipage} & \begin{minipage}[b]{\linewidth}\raggedright
\textbf{debris}
\end{minipage} \\
\midrule\noalign{}
\endhead
\bottomrule\noalign{}
\endlastfoot
train & 9,126 & 0 & 27,573 & 13,910 & 38,311 & 2,077 \\
validation & 10,208 & 0 & 2,042 & 0 & 5,698 & 0 \\
test & 1,225 & 0 & 11,322 & 1,535 & 5,285 & 0 \\
\end{longtable}

\begin{figure}[htbp]
\centering
\includegraphics[width=0.96\linewidth,keepaspectratio]{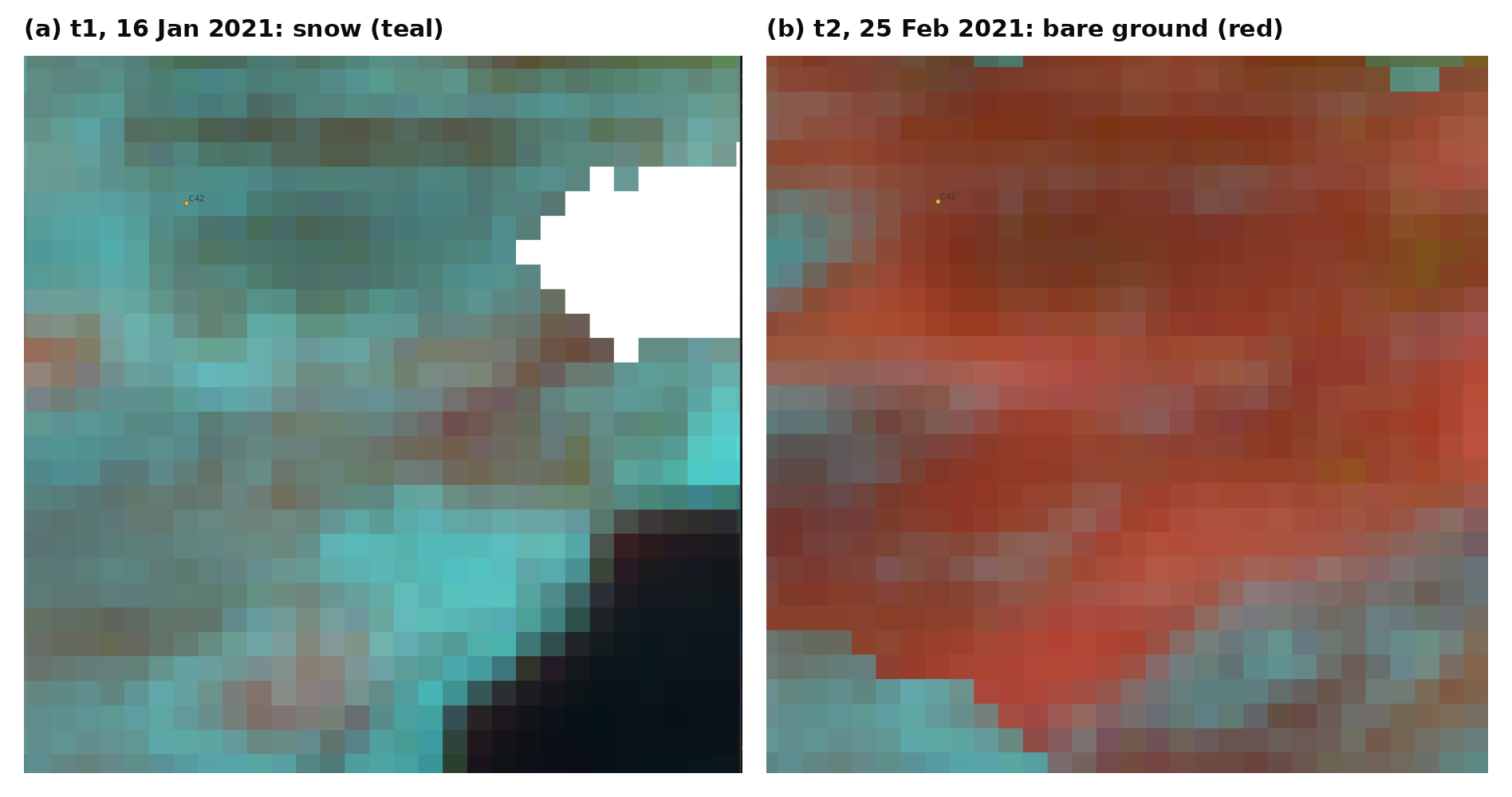}
\caption{Example of labeling from the blind change detection accuracy sample (C42, 1:1000, SWIR false color B12-B08-B04, each square is one 10 m pixel). (a) The green pixel (snow) at pre-event time t1. (b) Same pixel at post-event time t2 and is now red (bare ground), it has been labeled as change by the analyst. White regions are cloud/no data, and almost black regions are shadows of terrain where analysts have not selected the point to look at image.}
\end{figure}

For a change to be noted, the underlying material should be different for the two dates, like snow becoming bare ground. A change in terms of shadow and brightness alone does not count. Darkened, masked or points lying at the boundary between two materials for any of the two dates were not taken into consideration (for example, see Fig. 1). The Date Field refers to the date field for which the surface is available for polygon label observation. ``Both'' implies that the same surface is visible in both dates. ``t2'' means that the surface was snow covered, masked or in shadow in the pre-event image.

The reference event has three components. The source is located at 79.716431 E, 30.376403 N which is the detachment point of Ronti Peak. There are four checkpoints namely Ronti Gad, Rishiganga plant at around 15 km, confluence of Rishiganga-Dhauliganga and the Tapovan plant at 26 km. There are also three damage locations namely Rishiganga plant, Raini bridge and the Tapovan plant.

\section{Methods}\label{methods}

All thresholds, ranges, and rules are stated in a single configuration file and saved along with each run, along with input hashes, software versions, and random seeds. Figure 2 summarizes the six steps of the pipeline.

\begin{figure}[htbp]
\centering
\includegraphics[width=0.96\linewidth,keepaspectratio]{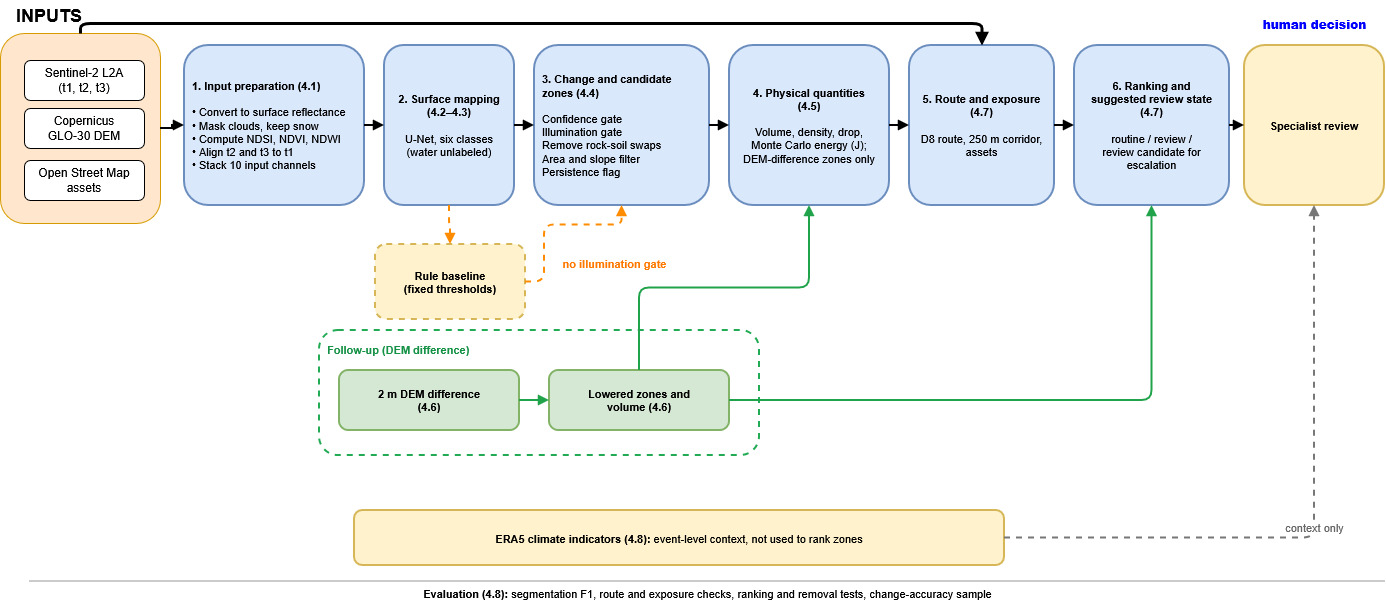}
\caption{Pipeline of screening process. Input goes through six stages and receives a ranked list of zones with the suggestion of review status; a particular specialist makes the escalated review decision.}
\end{figure}

In this figure is shown the pipeline of the screening process. As inputs, for this task, we used three time steps of Sentinel-2 L2A imagery (t1, t2, t3), the Copernicus GLO-30 DEM and the relevant OpenStreetMap assets. Pipeline includes six stages: (1) input preparation (section 4.1), (2) surface mapping with U-Net (4.2 and 4.3), (3) change and candidate zones (4.4), (4) physical quantities (4.5), (5) route and exposure (4.7) and (6) ranking with a suggested review status (4.7). Data on both DEM and asset are used at stage 5. The rule baseline (dashed orange) uses the fixed threshold values instead of U-Net and has no illumination gate. The difference-based DEM follow-up (green; section 4.6) computes lowered zones and their volumes and uses them for energy calculations (at stage 4) and ranking (stage 6). Calculation of energy is performed only for these zones. The ERA5 indicators (4.8, grey) are available only for events and are not used in ranking of zones. Software just suggests the review status. Escalated technical

\subsection{Input Preparation}\label{input-preparation}

The Sentinel-2 L2A DN (digital numbers) are transformed into surface reflectance R using equation (1), and from baseline 04.00 onwards, the ESA offsets by −1000:

\begin{quote}
R = (DN + O) / 10,000, O = −1000 if baseline ≥ 04.00, otherwise 0 (1)
\end{quote}

DN = 0 means that no data exist and R is limited to {[}0, 1.5{]}.

Confusion of snow and clouds may arise easily; thus, an extra condition must be applied for cloud masking procedure. Pixels of the cloud class of SCL {[}20{]} will be masked, except places of snow cover (NDSI \textgreater{} 0.4 and SWIR1 (B11) \textless{} 0.2). Cloud shadow of the SCL cloud class will be masked only within a range of 3 km and edges of the cloud will be extended by 2 pixels. Otherwise, the January image will have only 60\% of valid pixels left because of snow masking as cloud.

For classification we utilize three spectral indices calculated from B03 (green), B04 (red), B08 (nir) and B11 (swir1): NDSI {[}26{]}, NDVI {[}27{]} and NDWI {[}28{]}.

NDSI = (B03 − B11) / (B03 + B11) (2)

NDVI = (B08 − B04) / (B08 + B04) (3)

NDWI = (B03 − B08) / (B03 + B08) (4)

t2 and t3 co-registration with respect to t1 is performed via phase correlation {[}29{]}. Any residuals larger than 0.5 pixels are corrected. Image pairs whose residual shifts are greater than 2 pixels are then rejected (the t2 residual shift was 0.52 pixels and was therefore corrected). Aspect and slope are determined from the DEM using central differences {[}30{]}. The model input consists of a 10-band stack of bands B02, B03, B04, B08, B11, B12, NDWI, NDSI, NDVI and slope. The bands are normalized with their mean and standard deviation on the training strip.

\subsection{Geographic Split}\label{geographic-split}

Each strip is 64 pixels wide (640m). The 25\% eastern part of each strip is tagged as test area and one strip that is not used lies in between test and development data. The western 15\% of the scene is tagged as validation area and the rest of the strips constitute the training data. Patches in the training dataset can only come from training columns and thus labels which do not belong to training columns are masked out as ``ignore''. We shifted the validation area to the western edge of the screen since in the eastern part it was covering up the 2021 flow channel and hence there were no examples of debris in the training dataset.

\subsection{Surface Mapping}\label{surface-mapping}

The U-Net {[}10{]} (Figure 3) is trained to classify each pixel into one out of six categories (background, water, snow/ice, rock, soil and debris) in the three images, giving equal importance to the three strips. It is trained using a cross-entropy loss on the labelled pixels, where the class weights are calculated as follows:

wc = (1 / √nc) / {[} (1 / K) Σk (1 / √nk) {]} (5)

where nc is the number of training samples of class c, while K is the number of classes having training samples; the sum is taken over such classes only. The class having no training samples gets w = 0. Water has no labelled pixels in any strip, and hence its class weight is 0, making the model essentially a five-class classification network, and water is not considered for testing on the test strip.

Table 3: U-Net training and inference settings.

\begin{longtable}[]{@{}
  >{\raggedright\arraybackslash}p{(\columnwidth - 2\tabcolsep) * \real{0.4064}}
  >{\raggedright\arraybackslash}p{(\columnwidth - 2\tabcolsep) * \real{0.5936}}@{}}
\toprule\noalign{}
\begin{minipage}[b]{\linewidth}\raggedright
\textbf{Setting}
\end{minipage} & \begin{minipage}[b]{\linewidth}\raggedright
\textbf{Value}
\end{minipage} \\
\midrule\noalign{}
\endhead
\bottomrule\noalign{}
\endlastfoot
Architecture & U-Net, 32 base channels, depth 4 (four downsampling steps) \\
Input & 10 channels, 128 × 128 patches \\
Batch size, patches per epoch & 16, 800 \\
Optimizer & AdamW {[}31{]}, learning rate 10⁻³, weight decay 10⁻⁴ \\
Schedule & cosine annealing {[}32{]}, up to 30 epochs \\
Augmentation & random 90° rotations and horizontal flips \\
Model selection & lowest validation cross-entropy, patience 6 \\
Inference & sliding window, 256 px tiles, 32 px overlap, linear blending \\
Seeds & 0, 1, 2 (results given as mean +/- sd) \\
\end{longtable}

\begin{figure}[htbp]
\centering
\includegraphics[width=0.96\linewidth,keepaspectratio]{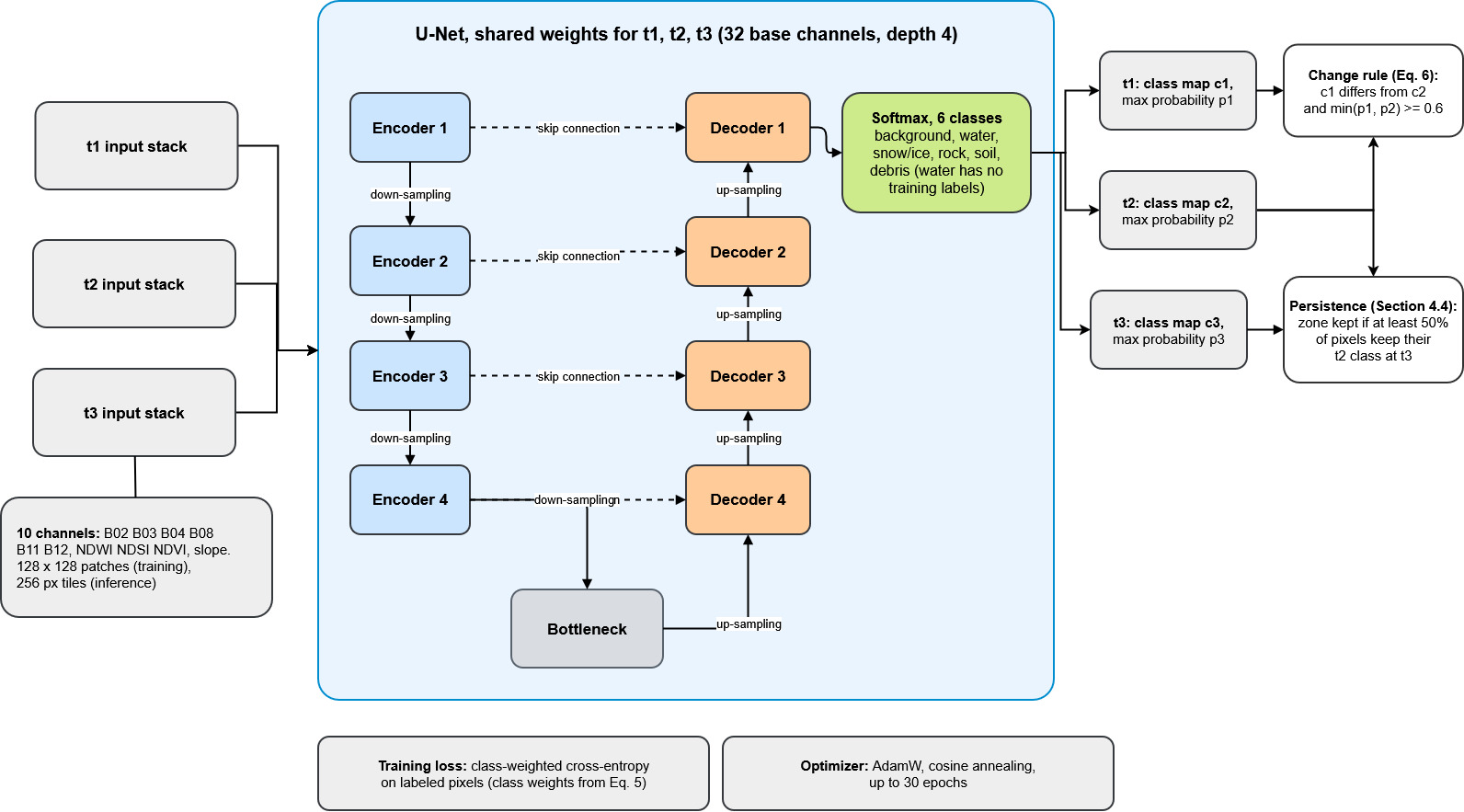}
\caption{A single U-Net with shared weights is used to perform surface mapping at t1, t2 and t3. The class maps are used to compute the change rule (Eq. 6) and the persistence check of a class in the map.}
\end{figure}

Surface mapping architecture (Section 4.3). The three images t1, t2 and t3 are treated by the U-Net using the same processing method, so that there is only one set of weights that is learned from the three dates. The 10-channel image B02, B03, B04, B08, B11, B12, NDWI, NDSI, NDVI and slope is divided into patches of 128x128 for training and 256 px for inference. The network uses 32 base channels and a depth of 4. This network is made up of 4 steps of an encoder which down samples, a bottleneck, and 4 decoder steps which upsample. There is a skip connection between each encoding step and its decoding step. The output layer of the network is softmax and produces 6 water classes. As there are no labels available for water, this model learns 5 classes, and for each of the dates in time, we store a 3D map of the class with the highest three probabilities and the maximum probability of each of the highest three classes (p1, p2, p3; c1, c2, c3). The evolution update rule of Eq. (6) uses t1 and t2 maps. The persistence test in Section 4.4 uses t2 and t3 maps. The training uses a weighted cross-entropy loss function where the class weights are determined by Eq. (5). AdamW optimizer with cosine annealing is used for optimization. Fig. 3 shows only the general interconnect structure. The number of channels per each interconnect level is not presented here.

The rules use fixed thresholds to assign each pixel into the same classes. Specifically, a pixel is classified as snow/ice if NDSI \textgreater{} 0.40 and green reflectance \textgreater{} 0.25, whereas a pixel is classified as water if NDWI \textgreater{} 0.15 and NIR \textless{} 0.15 and the pixel's slope is below 12°. Background pixels are classified to be vegetated if NDVI \textgreater{} 0.30. Within the bare pixels, a pixel is classified as rock if the pixel's slope ≥ 35°; soil if the pixel's slope \textless{} 20° and the 3×3 window's B11 3-SD \textless{} 0.025; and debris for all other cases. A pixel is flagged as changed if its rule classification is different between the two dates and the change vector magnitude M {[}14{]} for the common reflectance bands is greater than a scene specific threshold T = median(M) + 2.5 × 1.4826 × MAD(M). For these calculations, MAD refers to median absolute deviation. The value 1.4826 is used to scale the MAD such that it corresponds to the standard deviation of normally distributed data, {[}47{]} and the cutoff 2.5 is a standard value for a median based outlier criterion. {[}48{]}

\subsection{Change and Candidate Zones}\label{change-and-candidate-zones}

For each pixel, there are the class maps c1 and c2 and the softmax probabilities p1 and p2 of the two dates. The pixel is considered to be changed if

c1 ≠ c2 and min (p1, p2) ≥ 0.6 (6)

The elevation e and azimuth φ of the sun are extracted from the metadata of the image. Given the slope β and aspect α of a pixel, the cosine of the angle of solar incidence {[}33{]} is

cos i = cos θ × cos β + sin θ × sin β × cos (φ − α), θ = 90° − e (7)

A pixel is shadowed if cos i \textless{} 0.15 (shadow of the object itself or of grazing light) or if the terrain between the pixel and the sun obscures the path (cast shadow) up to 8 kilometers towards the sun. Those pixels are excluded which are shadowed on one of the dates and illuminated on the other. Also, we exclude those changes of rock to soil and vice versa, a condition that was imposed in view of the 324 such areas found in the first attempts.

Pixels modified through morphological cleaning operations (one pixel opening) are classified into objects through 8-connectivity operations. Objects that are less than 5,000 m² are deleted from the set. A dominant transition and source category are obtained for each of the selected objects. For instance, in the case of snow/ice and rock/soil categories, the requirement is that a median slope of at least 15° should be present. The persistence of zones depends on the presence of at least 50\% of the pixels in the zones belonging to the same category as t2 and t3. t3 is captured in March 2022 within the same season as t2.

The stability indices are saved in the same unit that they were derived from with their respective standard deviation. If the number of reference years exceeds 30, a normalized score from 0 to 1 (in the linear scale from the 10th to the 90th percentile of the reference values, cut off at the ends; Eq. (17)) is calculated for the indices. The minimum requirement of 30 reference years for the climate indices from Section 4.8 is met. Otherwise, there would be no score for this particular index at all. Missing values are never imputed. There is no combined index of instability. No stability index combination is used to create the instability index in the Chamoli test. Persistence is the only instability proof for a zone in the Chamoli test.

\subsection{Physical Quantities}\label{physical-quantities}

The mass m and the gravitational potential energy E of the source are expressed as follows:

m = ρ × V (8)

E = m × g × Δh = ρ × V × g × Δh (9)

where ρ stands for the density of the material (kg/m³), V stands for the volume (m³), g stands for the gravitational constant (9.80665 m/s²) and Δh stands for the weighted elevation drop along the path. Each volume, density and drop value is defined in terms of a particular distribution (uniform, normal, triangular or fixed) and then randomly sampled N = 20,000 times (Monte Carlo) in {[}34{]}. The median and 90\% confidence interval (5th - 95th percentile) of E are calculated for the volume, density and drop distributions. The contribution to the spread of the output made by each of the inputs is determined in terms of the normalized squared Spearman rank correlation coefficient {[}35{]} between the input and E. Intervals with a high/low ratio \textgreater{} 10 are indicated as ``inconclusive''.

The determination of the difference measured in a DEM and the determination of the available volume for the potential future release are two separate measurements. The volume difference determined using the DEM differencing for a given period does not give us any information about the thickness of the material that is potentially available for future releases. The ``loss volume'' determined using the DEM differencing is not taken into account when determining the available volume. Also, the area alone is never used to determine energy. Energy is determined for volumes in a thickness range which is independent. No value was given for optical zones in this case; therefore, no optical zone will have an energy value, and the optical zones will remain area change. Follow-up A (Section 4.6) is based on the determination of the measured lowering in the DEM difference as the volume of observed release, not as the volume of future release, which is going to take place, we use the energy index:

I = $\bar{\rho}$ × g × A × Δh (10)

where ``I'' is expressed in units of joules per meter of thickness of the source, so that I orders the zones according to energy order, assuming the same thickness of all zones. However, I is never expressed in energy. ``A'' represents the zone area and $\bar{\rho}$ is the mean density.

Gravitational potential energy is the physical screening parameter and as such it does not express failure probability, runout, impact force or loss. Thus, ranking according to energy or energy index is incapable of making any prediction of failure or consequences of a possible failure. The system also is able to give a probability of exceeding a user-specified screening value of energy. This feature was implemented but is not illustrated below because results of that analysis were not produced.

\subsection{DEM-Difference Volume and Its Error (Follow-up A)}\label{dem-difference-volume-and-its-error-follow-up-a}

Consider a zone with n pixels having pixel area ``a'' and elevation differences dzi (after event minus before event, excluding biases). The volume lost is given by

V = a × Σ \textbar dzi\textbar{} for n pixels of the zone, as given by Eq. (13) (11)

Noise in the stable terrain (no difference between the two DEM acquisitions) is estimated. The pixel is defined as stable if it has good values in both DEMs, is situated in flat terrain, and is not glacier or snow. The small lateral offsets in DEMs appear as large vertical errors in steep areas. Besides, the errors are not computed for the pixels classified as snow/ice in the maps of section 4.3. The stable terrain mask is a per-pixel mask in the 10-m grid. It is defined by the combination of three conditions for which the bias and the noise are calculated, and these two measures become meaningful only for the pixels where the mask is true. The noise is normalized median absolute deviation {[}17{]} of dz in the stable terrain:

NMAD = 1.4826 × median (\textbar dz − median(dz)\textbar) (12)

Pixel is referred to as lowered if

dz ≤ −max (5 m, 3 × NMAD) (13)

The volume error in the case of DEM error is a spatially correlated random field and hence makes use of effective number of independent samples. The area of the zone is determined using A = n × a and the correlation length as L = 100 m:

σV = NMAD × A / √neff, neff = max (1, A / (π × L²)) (14)

The source elevation, $\bar{z}$ is defined as the mass-weighted center of the lost column which is represented as $\bar{z}$ = Σ (zi + ½ di) × di / Σ di. The elevation of post-event DEM, zi is represented by middle of lost column zi + ½ di and di = \textbar dzi\textbar. The density for each zone will be in the range between 900 kg m⁻³ (glacier ice) to 2,700 kg m⁻³ (dense rock) {[}37{]}. It is important to note that from the elevation change, it is not possible to find out whether a change in mass was due to material or bulk change.

\subsection{Route, Exposure and Ranking}\label{route-exposure-and-ranking}

First, all depressions in the DEM are filled by morphological reconstruction {[}38{]}. Then each cell drains to the steepest one among its eight neighbouring cells (D8) {[}15{]}. A zone's route begins at the outlet of the zone, i.e. the cell of a zone with the maximum contributing area. Further downstream the route proceeds over up to 40 km in the zone. The reference elevation href is determined as the first route cell on the valley floor having a contributing area of no less than 2 km² and with the slope of no more than 8º. Δh = $\bar{z}$ -- href. If the route passes over the filled depression of depth larger than 30 m, the route is marked as a possible lake, dam or DEM error.

All cells that fall within 250m of the route constitute the screening corridor. In the Exposure category, all the inventory assets (settlements, bridges, hydropower) present within the screening corridor along with the distance from the route are provided. All zones/regions used in the assessment decision support system are ordered using a fixed key: (1) Proposed review status (review, review for escalation, routine); (2) Exposure number (more first); (3) Energy/energy index (large first); (4) Persistence (persistent first); (5) Alternative explanations (fewer first); (6) Area (large first).

It is not an operational warning or response system but its support. The zone will be considered Routine in case of lack of information, bad or short route being available. The zone will be classified as Review if the change is observed that is credible but unclear. The component will be regarded as a potential candidate for review escalation if there is some persistent change in the route that passes the exposed assets and at least two other signals of instability appear independently. For the Chamoli test case, however, persistence is the sole signal of instability (Section 4.4). No test zone can be considered a candidate for review escalation. In addition, the software cannot be put into the fourth state, namely to the state of escalated technical review. Software implementation of review candidate for escalation is absolutely different from the professional decision regarding escalated technical review. None of the automatic states will lead to any warnings, closure, evacuation or other engineering actions.

\subsection{Evaluation Protocol}\label{evaluation-protocol}

For every class c from the test strip, given the true positive TP, false positive FP and false negative FN {[}39{]}, we calculate the values of precision P, recall R (not the reflectance in Equation (1)) and F1:

P = TP / (TP + FP), R = TP / (TP + FN), F1 = 2 × P × R / (P + R) (15)

The Macro F1 value is the mean F1 value for all classes from the test strip label set -- background, snow/ice, rock and soil.

Suppose that there are N zones in total, and m of them are located within a given radius from a certain source. Then, the expected value of the best rank for ``m'' zones is:

E{[}best rank{]} = (N + 1) / (m + 1) (16)

We consider radio to 500 m, 1 km and 2 km.

The follow-up B uses the ranking on the zones which were downgraded in the follow-up A (Section 4.6). The ranking is done without energy, without persistence (instability), without exposure, and without route and in comparison, to simple baselines: largest area first, and largest volume or most certain change first. As for the optical zones, we were going to conduct the same test as described above. However, it is impossible to interpret the results of these tests if the source of the study is not detected (Section 5.4).

To test the route, we begin from the mapping of the route from a reliable source. If the route passes within 250 meters from the checkpoint, we assume that the route is reached. After this we compare the distances of the route with the distances that are written in the documents and damage of the sites mentioned in the documents to the actual corridor and inventory.

The metric used to evaluate the change accuracy is the Follow-up C. This is achieved by random sampling points in a stratum h of size Nh. We sample nh labelled points in a stratum and assign them a weight Wh = Nh / nh. We estimate the precision, recall and F1 of changes class using the weighted confusion counts {[}18{]}. There are cases where some of the strata had so little labelling that everyone agreed on it and hence bootstrapping would not make any sense in such situations. We report the uncertainty per stratum as exact (Clopper-Pearson) {[}40{]} 95\% confidence intervals of the change fraction in the respective stratum.

When evaluating the validation process for the algorithm for the test strip the following was done: we took a set of all pixels which are valid in both images (there are in total 678,791 pixels out of 3,180,000 in the scene which can be used). 4 strata were created using the change maps from the frozen U-Net algorithm run (seed 0) and the change map of the rule baseline. 70 points were randomly sampled at seed 29 with a minimum distance of 100m between them. These points came in 4 sizes: 15 (for both map variants), 15 (for the U-Net variant only), 15 (for the rule variant only) and 25 (for neither variant). Labels of C1 to C70 were assigned according to some random and therefore unknowable rules, see Section 3. All these three values of precision and recall were computed based on the evaluation of the same 70 points by three different seeds (0, 1 and 2) of U-Net. What we refer to as ``blind'' sample points is the particular selection of 70 randomly selected points in a stratum which were randomly labeled. Follow-ups were not blind because we developed them after learning about the optical issue (Section 7).

Follow-up D uses weather-based criteria. For each window of w days before the occurrence day, precipitation amounts and average temperature x {[}22{]} are calculated and compared with the corresponding values in the same window in the 30 years between 1991 and 2020. Using P10 and P90 to denote the 10th and 90th percentiles of the reference values, the criterion is

s = (x − P10) / (P90 − P10), restricted to {[}0, 1{]} (17)

\subsection{Implementation and Reproducibility}\label{implementation-and-reproducibility}

The pipeline is a Python module (Python 3.10+), with a command line interface. The pipeline is composed of several libraries such as NumPy {[}41{]}, SciPy {[}42{]}, scikit-image {[}43{]}, rasterio, Shapely and PyTorch {[}44{]}. It utilizes the pystac-client library for downloading the data from STAC catalogs. Experiments were conducted using the Kaggle notebooks with an NVIDIA Tesla T4 GPU. Order of the above methods is the same as the order of reading the results of the previous method. Thresholds and ranges are determined based on the single configuration described in Section 4. All layers are registered on one 10m UTM grid in zone 44N. In case when Planetary Computer DEM endpoint is unavailable, the same GLO-30 tile from AWS open data is used as explained in Section 5.8.

The configuration that is used in each screen is stored along with a manifest. The manifest file includes the SHA-256 hash value of all the input files, the hash value and other details of all the models that are used along with the version number of the library used and the random seeds that are used. The output of the run includes the class map and the change map (GeoTIFF file format), the zones and routes (GeoJSON and CSV formats) and the report. The output files (GeoJSON reference labels for the scenes, benchmark and event evaluation results) along with the follow up protocol and results are available on the code repository. All tasks are carried out in this sequence: installation of the package, gathering data, generating raster labels, conducting benchmark, collecting OpenStreetMap data, conducting screening and event assessment. The task of energy computation is repeated using the data set for the illustrative input values presented in Section 5.1 and the output data for the four corners compared to the corner examples\textquotesingle{} data set. An access to public data services over the internet is necessary to collect the data for the runs.

\section{Results}\label{results}

\subsection{Energy Calculation Check}\label{energy-calculation-check}

This was carried out using the energy layer for the following input values for Chamoli example: volume from 24 to 30 × 10⁶ m³, density 2,200 to 2,600 kg m⁻³ and drop 700 to 1,000 m. The above values were used in carrying out this verification test. Please note that the above volume falls within the range provided for the volume in published value of {[}3{]}, while the above values for density and drop volume were taken independent of other data. Independent uniform distributions have been used for each input across its entire range. 20,000 independent samples were taken for each input using the Monte Carlo method.

Table 4: Mass and energy for the input ranges given above (Eqs. (8) and (9)).

\begin{longtable}[]{@{}
  >{\raggedright\arraybackslash}p{(\columnwidth - 6\tabcolsep) * \real{0.4002}}
  >{\raggedright\arraybackslash}p{(\columnwidth - 6\tabcolsep) * \real{0.1753}}
  >{\raggedright\arraybackslash}p{(\columnwidth - 6\tabcolsep) * \real{0.2491}}
  >{\raggedright\arraybackslash}p{(\columnwidth - 6\tabcolsep) * \real{0.1754}}@{}}
\toprule\noalign{}
\begin{minipage}[b]{\linewidth}\raggedright
\textbf{Quantity}
\end{minipage} & \begin{minipage}[b]{\linewidth}\raggedright
\textbf{Low}
\end{minipage} & \begin{minipage}[b]{\linewidth}\raggedright
\textbf{Middle}
\end{minipage} & \begin{minipage}[b]{\linewidth}\raggedright
\textbf{High}
\end{minipage} \\
\midrule\noalign{}
\endhead
\bottomrule\noalign{}
\endlastfoot
Mass, corner values (kg) & 5.28 × 10¹⁰ & 6.48 × 10¹⁰ & 7.80 × 10¹⁰ \\
Energy, corner values (J) & 3.62 × 10¹⁴ & 5.40 × 10¹⁴ & 7.65 × 10¹⁴ \\
Energy, Monte Carlo 90\% interval (J) & 4.30 × 10¹⁴ & 5.37 × 10¹⁴ (median) & 6.61 × 10¹⁴ \\
\end{longtable}

The Monte Carlo range is much narrower than the range of the corner case, as it is unlikely for all three input variables to vary to an extreme level simultaneously. The reduction is equal to 64\% of the output variability (volume by 23\%, density by 13\%). Hence, the selection of a reference elevation is more crucial than the selection of density. The corner range is estimated to be about twice wide (3.62 to 7.65×10¹⁴ J). Thus, this agreement is a very weak argument. It is a sensitivity range, not a real energy of the event.

\subsection{Segmentation}\label{segmentation}

The average results for the held-out east strip test are represented by the rightmost bar in Fig. 4: U-Net performs better compared to rule-based technique (macro F1 of 0.648 ± 0.025 (sd) for 3 seeds vs. 0.519). The superiority comes from more accurate identification of soil (0.311 vs. 0.009) and background (0.982 vs. 0.713) pixels. Clearly, U-Net technique provides more significant improvements compared to the baseline for the soil and background classes. In addition, the class snow/ice is easily detected. This experiment cannot be conducted using water and debris samples because test strips are not available for these materials. Performance of U-Net on rock and soil samples is bad for both U-Net and baseline. U-Net shows 0.3 accuracy on rock class (0.301) and on soil class (0.311). Baseline shows 0.354 on rock class and only 0.009 on soil class. Because there are no test strip samples of water and debris, we cannot conduct an experiment with these samples on this split. U-Net had average accuracy of 0.740 ± 0.020 and baseline model had accuracy of 0.695.

\begin{figure}[htbp]
\centering
\includegraphics[width=0.96\linewidth,keepaspectratio]{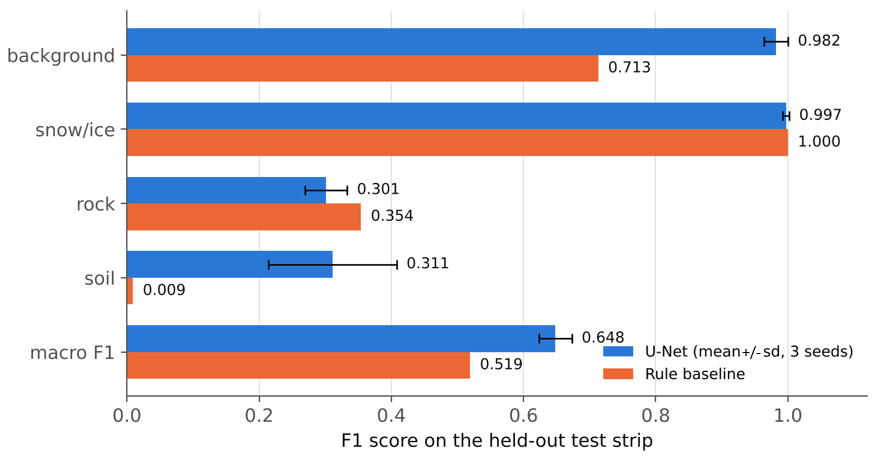}
\caption{Per-class F1 on the held-out test strip (1,225 background pixels, 11,322 snow/ice pixels, 1,535 rock pixels and 5,285 soil pixels). Blue bars show the mean and 1 std of the U-Net across the three seeds, orange bars show the rule baseline.}
\end{figure}

Our training with the particular GPU was not bit reproducible (using the same seed and the same settings gave a macro F1 of 0.747 in one trial and 0.617 in the other). We performed two runs of the benchmark using three seeds (0.643 +/- 0.078 and 0.705 +/- 0.091). What we present below are 10 trials of U-Net, which include 3 trials of the frozen benchmark, 3*2 trials of its two repeats and 1 trial of seed-0 (3+3+3+1). In all 10 cases the mean value for seeds is above the baseline. The highest F1-scores were reached for soil (F1 from 0.015 to 0.755) and rock (from 0.170 to 0.492). Thus, the mean values over seeds are the only numbers that should be reported (see Fig. 5). Seed 2, frozen trial, best checkpoint for epoch-1. It follows that the validation loss is a poor selection criterion for rock and debris since there is none in the validation strip.

\begin{figure}[htbp]
\centering
\includegraphics[width=0.96\linewidth,keepaspectratio]{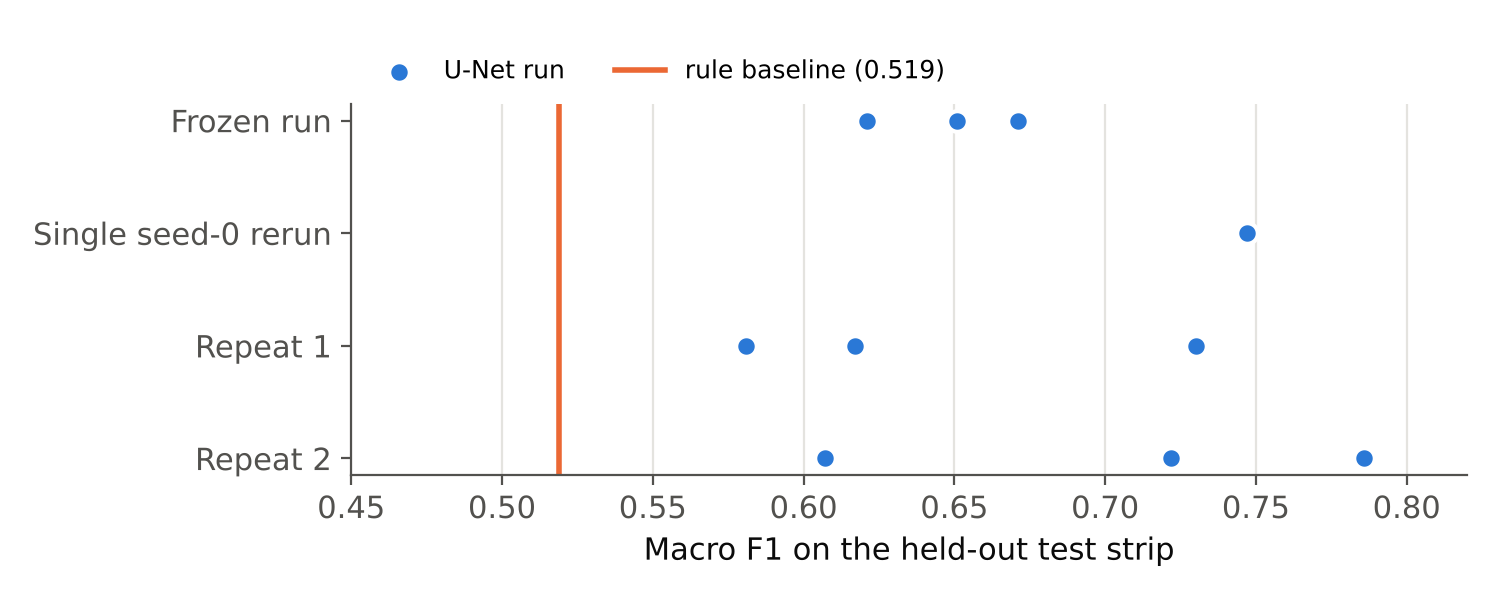}
\caption{F1 Macro scores from all runs of U-Net (a total of 10 runs, including three seeds from the frozen benchmark as well as three seeds for each repeat of download source change, two repeats in total). Both the repeats were run under the same labeling and settings after the download source for DEM data was changed to the same GLO-30 data product. The orange line is the baseline for the rule method.}
\end{figure}

\subsection{Change Mapping}\label{change-mapping}

The frozen U-Net pass (seed 0) yielded 302 zones of interest from which 24 had been rejected according to slope criteria to yield the map illustrated in Figure 6. The illumination gate removed 9.15 km² of class change, the rule of rock and soil class exchanges removed 3.17 km². The most extensive areas of change were snow/ice and debris (2.86 km²), background and soil (1.93 km²), snow/ice and soil (1.67 km²), soil and debris (0.99 km²), snow/ice and rock (0.68 km²). Corresponding zones of the 2021 debris in Rishiganga and Dhauliganga Valley had changes in debris classes. The number of zones in a changed area to debris was highly variable depending on seeds and passes (126 to 371 for the U-Net) and 434 for the rule baseline without an illumination gate.

\begin{figure}[htbp]
\centering
\includegraphics[width=0.96\linewidth,keepaspectratio]{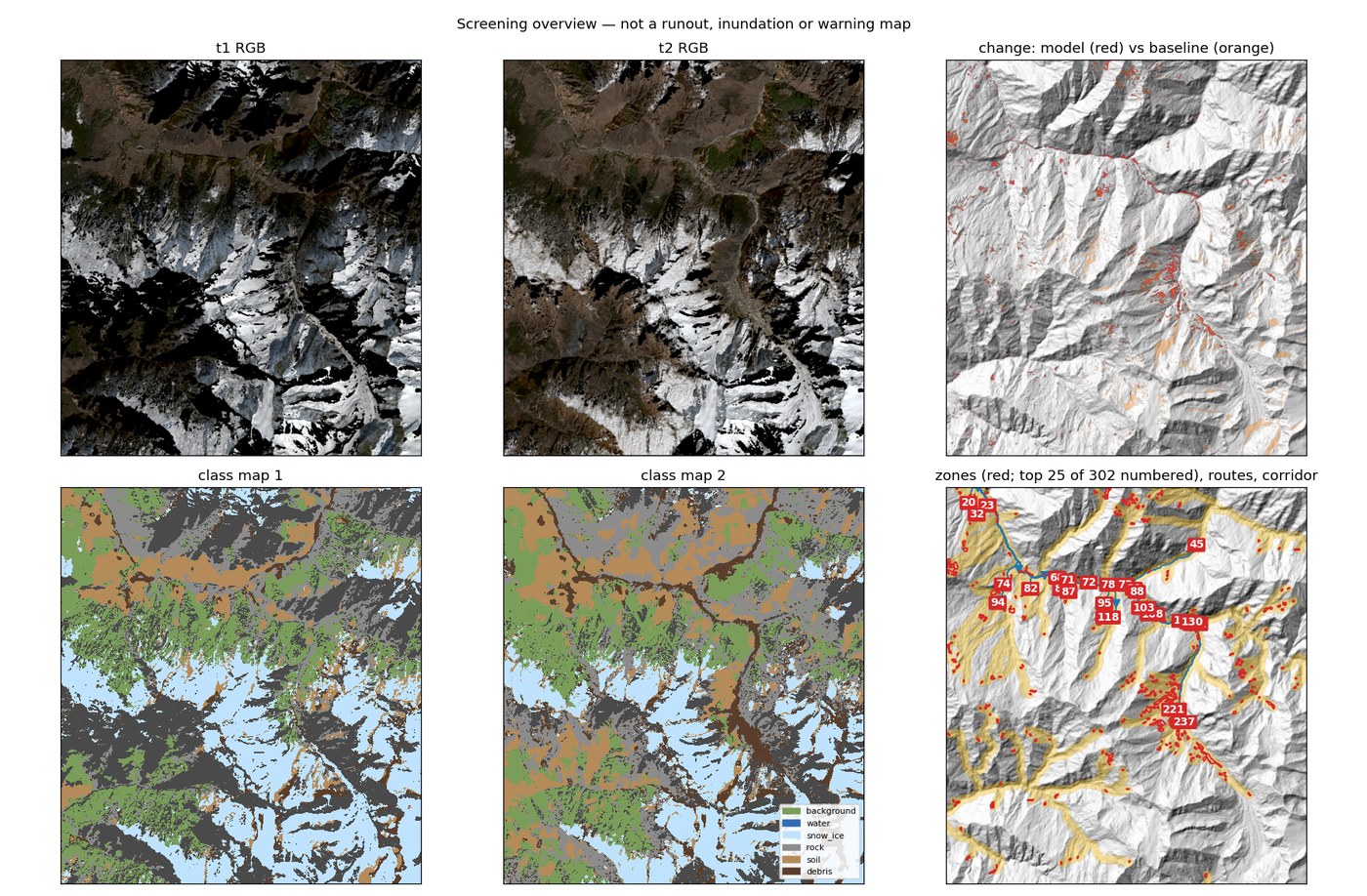}
\caption{Inspection of output from the run of the U-Net trained with the frozen architecture (seed 0): True color images at t1 (January 16, 2021) and t2 (February 25, 2021), upper left and center panels; Upper right panel: Changed pixels for the U-Net (red) and rule baseline (orange) on a hillshade. Lower left and center panels: Class maps of the U-Net for both time periods. Candidate zones (red, top 25 out of 302 sorted by review order) along with their pathways (blue) and 250 m corridors (yellow), lower right panel.}
\end{figure}

\subsection{Event Evaluation with Optical Change}\label{event-evaluation-with-optical-change}

The optical change maps did not show this feature. Examining the change maps within 1 km around this identified feature, the face of this feature has a slope of 39° and an aspect of 44° (NE). 73\% of the area was in the shadow during t1, 47\% in the shadow at t2 and 47\% in the shadow on both dates. 26\% of the area was eliminated by the illumination gate since its shadow status changed between the two dates while only 27\% of the area was illuminated on both dates.

Table 5: Optical changes around the identified source, by run.

\begin{longtable}[]{@{}
  >{\raggedright\arraybackslash}p{(\columnwidth - 6\tabcolsep) * \real{0.1779}}
  >{\raggedright\arraybackslash}p{(\columnwidth - 6\tabcolsep) * \real{0.2847}}
  >{\raggedright\arraybackslash}p{(\columnwidth - 6\tabcolsep) * \real{0.1988}}
  >{\raggedright\arraybackslash}p{(\columnwidth - 6\tabcolsep) * \real{0.3387}}@{}}
\toprule\noalign{}
\begin{minipage}[b]{\linewidth}\raggedright
\textbf{Run}
\end{minipage} & \begin{minipage}[b]{\linewidth}\raggedright
\textbf{Changed pixels within 1 km, before zone rules}
\end{minipage} & \begin{minipage}[b]{\linewidth}\raggedright
\textbf{Pixels in zones}
\end{minipage} & \begin{minipage}[b]{\linewidth}\raggedright
\textbf{Zones within 500 m / 1 km / 2 km}
\end{minipage} \\
\midrule\noalign{}
\endhead
\bottomrule\noalign{}
\endlastfoot
U-Net seed 0 & 112 & 0 & 0 / 0 / 1 \\
U-Net seed 1 & 0 & 0 & 0 / 0 / 0 \\
U-Net seed 2 & 274 & 199 & 0 / 1 / 1 \\
Rule baseline & 763 & 150 & 0 / 1 / 4 \\
\end{longtable}

Table 5: None of the runs has any zones mapped within 500 meters of the source. The zones that lie at distances between 500 meters and 2 kilometers from the epicenter change from debris or soil to snow/ice, hence, gain snow. Detachment results in loss of snow/ice and appearance of new rock. \textquotesingle Zones\textquotesingle{} described here are not source detections and therefore, optical ranking and removal analysis on them are yet to be done. For the sake of completeness, the nearest zone is ranked 6th of 131 in the seed 2 run and 12th of 434 for the rule baseline run under full ranking.

The route mapped from the documented source matches the documented flow path. It is the same for all runs since it only uses the DEM data and reaches all the four published checkpoints in order (Table 6, Fig. 7).

Table 6: Route distance to the documented checkpoints (published distances from Shugar et al. {[}3{]}).

\begin{longtable}[]{@{}
  >{\raggedright\arraybackslash}p{(\columnwidth - 6\tabcolsep) * \real{0.4214}}
  >{\raggedright\arraybackslash}p{(\columnwidth - 6\tabcolsep) * \real{0.1724}}
  >{\raggedright\arraybackslash}p{(\columnwidth - 6\tabcolsep) * \real{0.2247}}
  >{\raggedright\arraybackslash}p{(\columnwidth - 6\tabcolsep) * \real{0.1815}}@{}}
\toprule\noalign{}
\begin{minipage}[b]{\linewidth}\raggedright
\textbf{Checkpoint}
\end{minipage} & \begin{minipage}[b]{\linewidth}\raggedright
\textbf{Closest approach (m)}
\end{minipage} & \begin{minipage}[b]{\linewidth}\raggedright
\textbf{Distance along route (km)}
\end{minipage} & \begin{minipage}[b]{\linewidth}\raggedright
\textbf{Published distance (km)}
\end{minipage} \\
\midrule\noalign{}
\endhead
\bottomrule\noalign{}
\endlastfoot
Ronti Gad & 78 & 2.4 & not given \\
Rishiganga plant & 28 & 15.6 & about 15 \\
Rishiganga-Dhauliganga confluence & 51 & 17.1 & not given \\
Tapovan plant & 28 & 23.5 & 26 \\
\end{longtable}

The alignment passes through a depression of 34 meters in the gorge, which is a lake, dam, or an artifact on DEM, and the issue was not analyzed for the purposes of Section 4.7. The path is 2.5 kilometers or roughly 10\% shorter than the distance to Tapovan. This finding is in accordance with one long line of D8 direction over a meander bend, followed by the main flow.

\begin{figure}[htbp]
\centering
\includegraphics[width=0.96\linewidth,keepaspectratio]{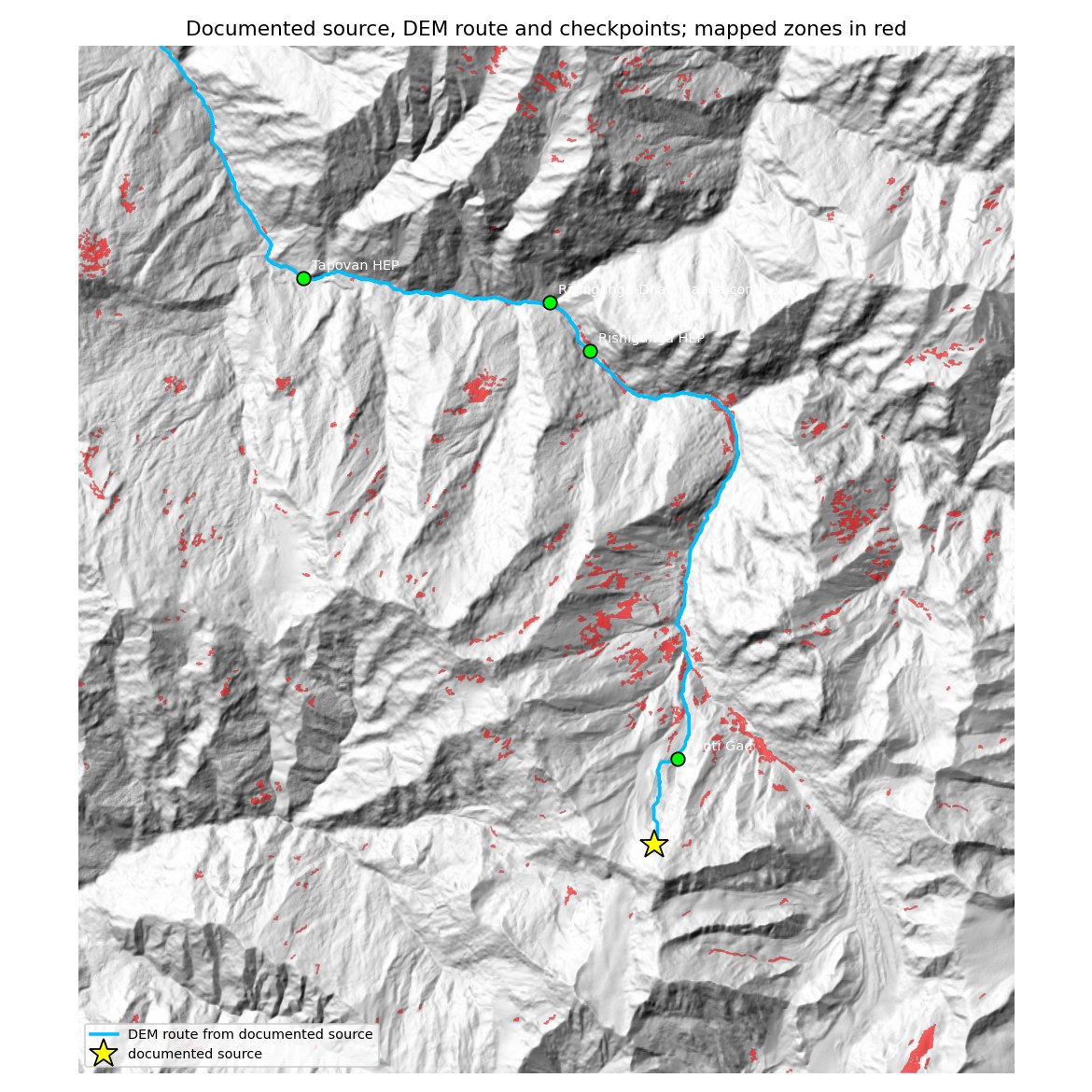}
\caption{Route derived from source (blue) overlaying the hillshade with source (star) and four checkpoints (green). The baseline used is the same for all zones in (red).}
\end{figure}

There is an incomplete inventory of the exposed assets in the corridor. There were six assets from OpenStreetMap {[}24{]} that were found to be present in the corridor including one destroyed bridge at Raini (16.6 km), a village (Raini, 16.9 km), a Joshimath-Malari Road bridge (18.0 km), an unidentified bridge (24.3 km), village Garghaniya, village (25.0 km) and an unidentified bridge (25.2 km). All three of the documented damaged assets lie in the corridor (3/3). But of the three, only the damaged bridge at Raini has been documented in the damaged assets (1/3). There was no hydropower facilities found in this map using OpenStreetMap.

\subsection{Follow-ups A and B: Detection and Ranking from Elevation Change}\label{follow-ups-a-and-b-detection-and-ranking-from-elevation-change}

DEM Difference also helped to establish the cause of error, i.e., the 2 m difference (errors) turned out to include 67\% of the study region {[}8{]}. As far as the 576,851 stable terrain pixels discussed in Section 4.6 are concerned, vertical bias was equal to −0.31 m and NMAD -- 3.24 m. Thus, the threshold for lowering was calculated to be 9.73 m (3 × NMAD, where the value of NMAD was rounded to 3.24 m).

There were 552 lowered regions (see Figure 8). The bottom three regions are situated within 500 m of the known source and are characterized by 183 m of lowering. Elevation change is not dependent on illumination; that is why several lowered regions were detected, which had escaped the notice of winter optical imagery.

\begin{figure}[htbp]
\centering
\includegraphics[width=0.96\linewidth,keepaspectratio]{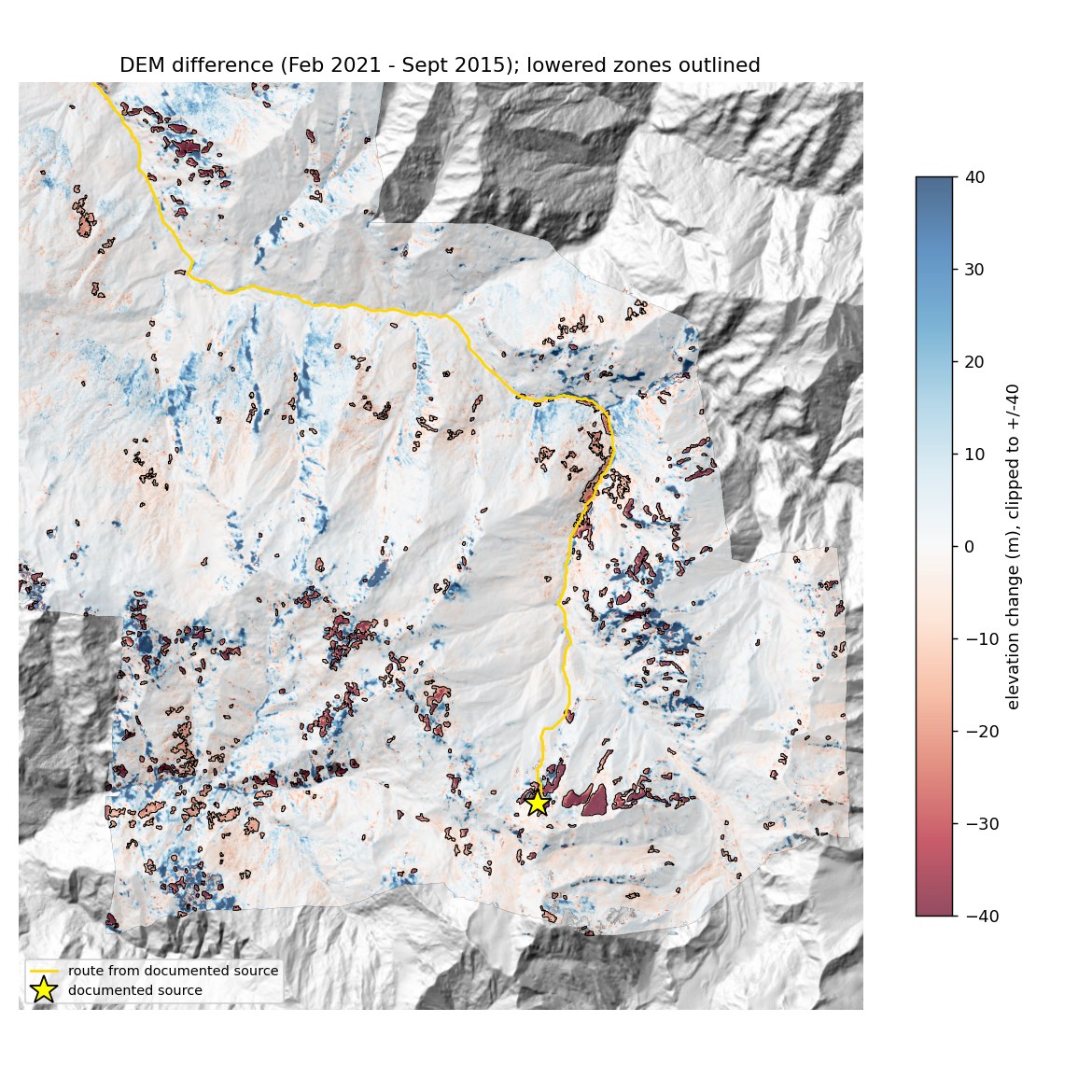}
\caption{Change in elevation from September 2015 to February 2021, 2 m DEMs resampled to 10 m (red=lowering, blue=raising; clipped at +/- 40 m). The 552 areas where lowering occurs are marked, and the source is pointed out using a star, and the path taken is shown (yellow). The area that has maximum lowering is on the east side of the star.}
\end{figure}

The declared rule just picks up the largest area within a radius of 1 km from any point which is well documented elsewhere. In this case, it picked up an area of 291,000 m² with a lost volume of 1.40 × 10⁷ m³ (95\% CI: 1.33 × 10⁷ m³ to 1.46 × 10⁷ m³). It is 0.52 times the volume of the well-documented detachment volume of 2.69 × 10⁷ m³ {[}3{]}. The well-documented detachment volume is not within the range of this volume. The mass-weighted elevation of this zone is 4,963 m (the well-documented detachment is at about 5,500 m), the drop is 1,394 m, and the energy is 3.44 × 10¹⁴ J (90\%: 1.89 × 10¹⁴ J to 4.98 × 10¹⁴J).

The post-hoc consideration of the zone 1.5 km to the east of the reported point in the same face (5,414 m, 307,600 m², average lowering 85 m) has 2.61 × 10⁷ m³ (97\% of the published volume), and energy 7.5 × 10¹⁴ J (90\%: 4.1 to 10.9 × 10¹⁴ J). It is well within the range ``corner'' of 3.6 to 7.7 × 10¹⁴ J estimation presented in section 5.1. With mid-range density 1,800 kg m⁻³, the height of fall is 1,630 m (rather than 700 to 1,000 m as estimated in Section 5.1). Hence, the agreement is achieved due to increased height of fall and decreased density in comparison with reference data. The primary scar in this zone is found at its west edge, and its coordinates along with those of the source feature are reported in {[}3{]}. The published volume was taken from the same DEMs as the published volume; hence, the agreement is not an independent check. As usual, the path from the stated source zone passes all four checkpoints in the same order (Rishiganga plant at 15.1 km, Tapovan plant at 23.1 km).

Energy-conserved ranking of the pipeline resulted in its position of 14th out of 552 pipelines regarding best zone within 500 m and 1 km and 13th best zone within 2 km.Randomization of ordering would yield 138, 50 and 23 (Eq. (16)). Upper 2.5\% of the observations analyzed in the post hoc test are presented here. The zone under consideration in this graph is located 1.5 km from the depicted point. Hence, it can just get into the 2 km range from the depicted point. The removal of exposure or route factor resulted in 2nd position of the best zone (1st within 2 km) -- as with ranking by area/volume only.

\begin{figure}[htbp]
\centering
\includegraphics[width=0.96\linewidth,keepaspectratio]{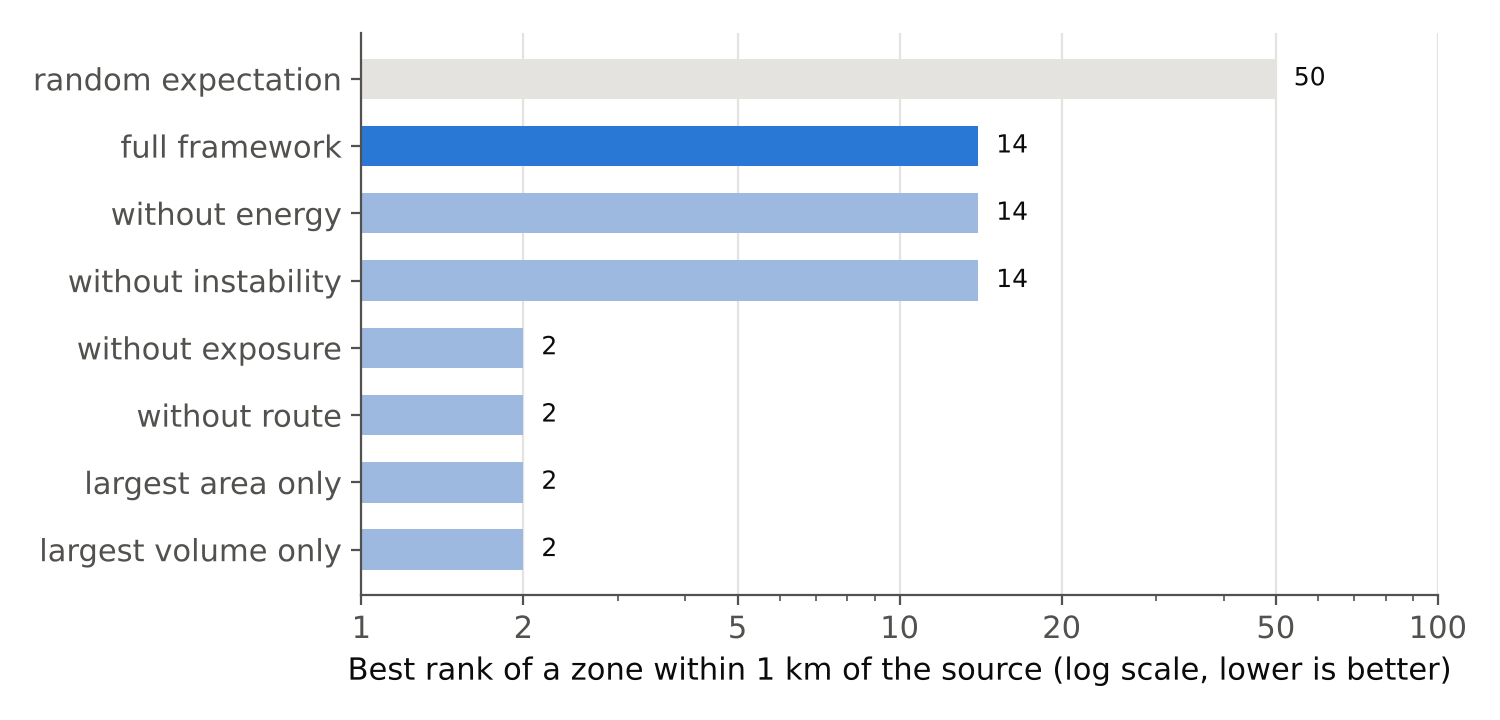}
\caption{Lowest ranked zone of 1km buffer to source of 552 zones for complete pipeline and also with removal of a single component of the pipeline. Two simple baseline results are also presented (log scale; best is lowest rank). Expected rank from a random ranking is shown in gray.}
\end{figure}

Thirteen most ranked zones in 1 km are small and lowered zones in the Rishiganga gorge, and their corridors hold seven assets compared to six for the best-known source route (section 5.4). We have not traced the difference in individual asset scores. This might be due to the location of routes or some assets near the end of the 250 m corridor. Exposure is counted first before the energy so that a difference of one asset counts for more than a difference of 30 in the energy. For this reason, the removal of energy makes no difference because energy is just used to break a tie in exposure score. Note that as stated previously, the removal of instability makes no difference because persistence is not turned on for the DEM zones

\subsection{Follow-up C: Change Accuracy}\label{follow-up-c-change-accuracy}

Among the 70 blind points, 43 or 61\% remained un-assessed due to terrain shading that prevented observation on either or both dates (Table 7).

Table 7: Sample of blind change accuracy, by stratum.

\begin{longtable}[]{@{}
  >{\raggedright\arraybackslash}p{(\columnwidth - 6\tabcolsep) * \real{0.2952}}
  >{\raggedright\arraybackslash}p{(\columnwidth - 6\tabcolsep) * \real{0.2458}}
  >{\raggedright\arraybackslash}p{(\columnwidth - 6\tabcolsep) * \real{0.2451}}
  >{\raggedright\arraybackslash}p{(\columnwidth - 6\tabcolsep) * \real{0.2139}}@{}}
\toprule\noalign{}
\begin{minipage}[b]{\linewidth}\raggedright
\textbf{Stratum (pixels)}
\end{minipage} & \begin{minipage}[b]{\linewidth}\raggedright
\textbf{Labeled / sampled}
\end{minipage} & \begin{minipage}[b]{\linewidth}\raggedright
\textbf{Change / no change}
\end{minipage} & \begin{minipage}[b]{\linewidth}\raggedright
\textbf{95\% exact interval, share changed}
\end{minipage} \\
\midrule\noalign{}
\endhead
\bottomrule\noalign{}
\endlastfoot
both maps changed (3,487) & 10 / 15 & 10 / 0 & 0.69 to 1.00 \\
U-Net only (3,692) & 4 / 15 & 0 / 4 & 0.00 to 0.60 \\
rule baseline only (70,276) & 2 / 15 & 2 / 0 & 0.16 to 1.00 \\
neither (601,336) & 11 / 25 & 0 / 11 & 0.00 to 0.28 \\
\end{longtable}

All labelled transitions with visible ground were instances of snow transitioning to bare ground. While the baseline model classified all changes as true changes, the four points selected by the U-Net models were false positives. Point estimates of precision and recall of ``change'' using area-weighted averaging were 0.49, 0.05 for U-Net seed 0, 1.00, 0.51 for seed 1, and 1.00, 0.04 for seed 2. All were evaluated based on 70 points and strata were created from the seed 0 change map. The baseline was rated at 1.00 and 1.00. These ratings are not very precise. The stratum of the baseline only contributing to the recall is determined by two labels (rest on two labels). In the case of 0 out of 11 samples remaining unchanged, there was still a possibility of 28\% change in the image, which would considerably reduce recall in all these cases. There is no way of verifying most of the change areas in the image from a distance of 10m in winter, and the gates in the U-Net pipeline for this section of the image have been set to mask the rare change cases detected.

\subsection{Follow-up D: Weather Before the Event}\label{follow-up-d-weather-before-the-event}

This analysis was performed using the ERA5 {[}22{]} cell at the source point (elevation of 5,297 m returned by the Open-Meteo service that can be different from the native ERA5 cell). During the 30 days ending with February 6, 2021, 19.3 mm of precipitation was recorded in comparison with a median value of 115.7 mm over the 19 reference years. The mean temperature was 2.16 °C warmer than the 30-year mean. The precipitation was less than the 10th percentile of the reference years and the temperature was higher than the 90th percentile of the reference years. Consequently, both scores are at the end of the truncated distribution of Eq. (17), that is 0.00 and 1.00. Scores for the last 1 to 7 days were close to normal (Fig. 10). Winter mean (Dec to Feb) temperature for the region in the past 29 years is increasing by +0.05 °C/decade. Temperature anomaly for an individual 31 km grid cell does not qualify the larger scene and anomaly alone does not establish cause.

\begin{figure}[htbp]
\centering
\includegraphics[width=0.96\linewidth,keepaspectratio]{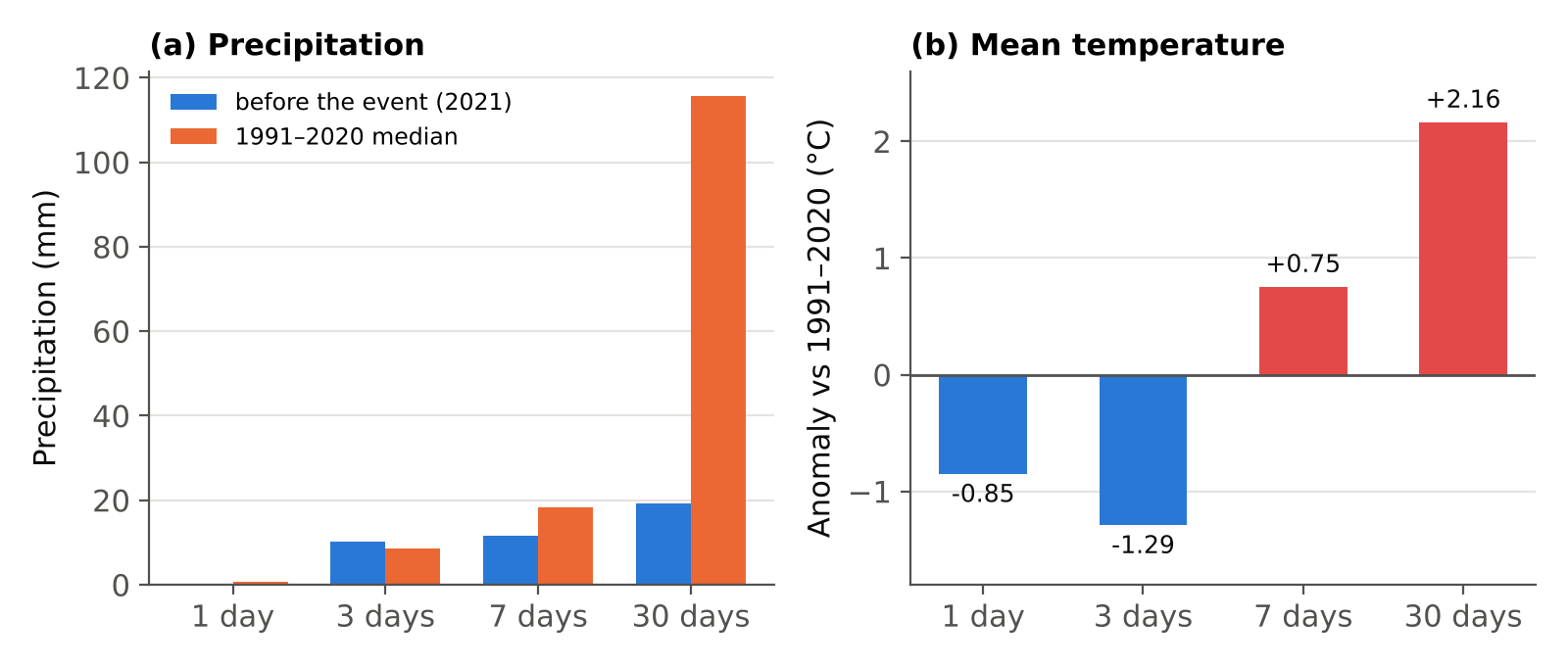}
\caption{Precipitation sums (a) and temperature anomalies (b) for ERA5 computed using windows of 1, 3, 7, and 30 days until 6 Feb 2021, averaged over the source grid box compared to the corresponding data for the period 1991-2020.}
\end{figure}

\subsection{Development Log: Failures and Their Fixes}\label{development-log-failures-and-their-fixes}

Apart from the last one, all other changes in Table 8 have been made based on preliminary results from the scene in question and its runtime compared to the frozen scene has not been documented separately. The development and event scenes are similar and therefore all changes have been done based on data from the event scene. Consequently, the main test has been a post-test rather than an independent test. Segmentation test (refer Section 4.2) is the only test which has been geographically different from the training and validation tests. The last change concerns the source of DEM download.Table 8: Observations in early runs and the changes made.

\begin{longtable}[]{@{}
  >{\raggedright\arraybackslash}p{(\columnwidth - 2\tabcolsep) * \real{0.5256}}
  >{\raggedright\arraybackslash}p{(\columnwidth - 2\tabcolsep) * \real{0.4744}}@{}}
\toprule\noalign{}
\begin{minipage}[b]{\linewidth}\raggedright
\textbf{Observation in early runs}
\end{minipage} & \begin{minipage}[b]{\linewidth}\raggedright
\textbf{Change made}
\end{minipage} \\
\midrule\noalign{}
\endhead
\bottomrule\noalign{}
\endlastfoot
Clouds within the default cloud mask are stripped out of the image. This is what led to the January image having 60\% valid pixels. & cloud mask adjusted to preserve snow cover (Section 4.1) \\
The February 2020 picture taken before the event had 1,400 zones that were primarily annual fluctuations of snow. & \\
Low winter sun signal detected only on one day when slopes were illuminated. & \\
October t3 "confirmed" that there was seasonal snow melt & pre-event image relocated to the same winter (Jan. 16, 2021) \\
The eastern validation strip included all debris zones & illumination gate, Equation (7) \\
Validation macro F1 was unreliable when using it on an image that contained only two or three classes. & \\
Rock-soil swap zones resulted in 324 zones & t3 (March 2022) and seasonal persistence \\
Test strip did not contain rock or soil zones & validation was shifted to the western boundary \\
Planetary Computer DEM endpoint stopped working in November 2026. & model selection via cross-entropy validation \\
\end{longtable}

\section{Discussion}\label{discussion}

Routing did work after defining the source, but the energy test worked only partially. The D8 single flow line, which is generated using the documented source based on 30 m free elevation data, is flowing through all the four documented checkpoints in the right sequence. The area of 250 m length is completely enclosed by the recently built seismic storage structure. This 250 m area contains all three buildings that were damaged. Public inventory contains data about just one building being damaged. This is a test for routing and exposure, not for sources. The route is from the documented source. The energy check in Section 5.1 is even more ineffective. In essence, it is just repeating the calculations made in Section 5.1 using the same input data, which means that it verifies that Section 5.1 uses the correct equations (8) and (9) for the energy check. Unfortunately, the results do not provide sufficient evidence that allows us to prove the correctness of the energy. The uncertainty is carried out in the process of calculation, and it is not a separate step. But according to the declared volume criterion, the layer regains only 0.52 of the published volume (Section 5.5). This zone (7.5 × 10¹⁴ J) found afterwards within the limits of the corner of Section 5.1 is just an observation.

The weakest element in the process of the source identification for wintertime is the optical detection of the source. The source is the steep northeast facing area, which is shaded in January and partly illuminated in February. Due to the low sun angle causing false changes in other parts of the image (which eliminated 9.15 km² of the changed area), we have included the illumination gate in the process. Detection of the change at the particular source was hampered by elimination of pixels with both change and signal while eliminating false alarms. Other north facing sources could encounter the same problem with winter optical images, although it will remain a hypothesis until more evidence becomes available. It should be noted that elevation change is independent of illumination and apparently identifies the source right away {[}8{]}. It should be noted that this kind of detection is based on post-event image, meaning that it can detect the release only after it occurs. So, this type of detection does not provide warning of an impending release.

The surface classification did not translate into better change detection. Although the U-Net in this case outperformed the rules on per-class F1 scores, the rules scored better on the visible ground class for instance detecting the change from snow to ground that the U-Net did not. The confidence and illumination gates are implemented on top of the change detected by the U-Net, and not on the classes. Thus, when only looking at the segmentation scores, one overestimates the U-Net's ability to detect change.

The ranking rule needs to be redesigned. The current ranking rule first orders by the assets' exposure and then by their energy. As a result, a difference of one asset can outweigh a difference of 30 in energy. Exposure counts from a course, incomplete inventory are a weak primary key. A method that combines energy with exposure (or sorts by energy within exposure bins) should report the used rule before testing the next case. We did not change the method here for the reason that tuning a method for its own test case would be stale.

Both methods failed to separate between rock and soil at 10 m in winter (F1 of at most 0.35). As the cost of not detecting a failure that goes from rock to soil but exposes fresh rock on a soil-like surface is not detected in optical imagery, this issue is ignored for now. In a winter scenario at 10 m, the performance of both methods to discriminate between rock and soil was weak (F1 of at most 0.35). Therefore, these results will be ignored for the scope of this report.

The public inventory {[}24{]} did not include the two most damaged structures of the ones assessed. For hydropower and other critical exposed assets, the asset data must come from an authoritative register for exposure screening to be as complete as possible.

\section{Limitations}\label{limitations}

The following limitations are listed in order of impact on the findings. Where the paper deals with a limitation, a bullet point leads to the corresponding section.

\begin{itemize}
\item
  This is a single event, single scene and single geographic split. The settings were tuned on the same scene that contains the event (Section 5.8). This test is a retrospective test and therefore not independent. No non-event area exists for this experiment. Future work will test on a second event as well as on non-event areas (Section 8)
\item
  Winter optical change detection failed to identify the source of change as the pixels that were removed by the illumination gate were in shadow where change would show up best. One event is not enough to get an idea of how frequently this occurs for other sources. The shadow-insensitive detection using SAR images or multi-date DEMs is the first issue of future work.
\item
  The successful source detection here uses post-event DEMs to identify a burst after it has occurred. Therefore, this pipeline as tested does not provide early warning of a future outburst. These follow-up experiments were designed after the optical failure had become known. Thus, these experiments are not blind tests (Section 4.8).
\item
  The ranking rule used first exposure count and then energy. Therefore, a difference of one asset outweighed a 30-fold difference in energy. No change to the rule for this case in Section 6. The rule that is proposed in Section 8 to declare a priori whether a test is to weigh energy or exposure to change the behavior of the test would be appropriate for this case as well.
\item
  This energy was not compared with an independent published estimate for Chamoli. Dependence on a number of assumptions including reference elevation, density and drop ranges for which no independent information is available (Section 5.1 and 5.5). The declared volume rule was implemented to recover 0.52 of the declared volume, and the best energy agreement was negotiated for one of the zones after the fact.
\item
  One person created all the labels on the 10 m imagery and there is no second person to check them against. Furthermore, water has no labels on any of the strips so effectively the model is being trained on 5 classes (i.e. land cover types) on the 10 m imagery. This method was not tested on the Debris dataset because it has no test-strip labels, and the rock test set is very small (1,535 pixels). For the rock test set, pixels. Pixels from the same polygon are not independent and thus the spread over seeds underestimates the uncertainty of the pixel-level F1 scores.
\item
  The validation strip has no rock or debris, so model selection cannot favor those classes. For seed 2 of the frozen run, the epoch-1 checkpoint was selected as best. GPU training is not bit-reproducible, so we report means over seeds and the range over ten runs (Section 5.2), and the number of candidate zones varied from 126 to 371 between U-Net runs. The main change maps and candidate zones, and the strata of the point sample, come from the seed 0 run.
\item
  The change-accuracy sample has very few labels in the strata that drive recall, and 61\% of the points could not be judged. Its precision and recall values are therefore qualitative (Section 5.6).
\item
  Slope, aspect and shadow have been derived from the 30 m GLO-30 DEM {[}21{]} resampled to 10 m to smooth narrow ridges and gullies. The co-registration was done by means of a single global sub-pixel shift only, so local misregistration is not corrected. The boundaries of the study area were drawn approximately and not matched to the boundaries of the source polygon described in Section 3.
\item
  The 2 m DEM difference {[}8{]} covers the time span from 2015 until 2021. Thus, five years of glaciers' changes as well as of their pre-failure movements and of the snow cover are recorded. The same DEMs were used to compute the published volume {[}3{]}. Thus, agreement with the published volume {[}3{]} is not independent to check.
\item
  OpenStreetMap {[}24{]} in 2026, five years after the event, has no hydropower features in the database. Hence the two most damaged structures in the exposure list were not included in the database. The exposure counts depend on the same list as this inventory. Why do the corridors of the 13 DEM zones ranked highest pass seven assets and six assets on the same way along the routes as documented in Section 5.5.
\item
  The only form of instability evidence is persistence. There is no evidence that any zone can reach a Review Candidate state (Section 4.7). The ERA5 climate information provided are for a single event and a single grid point (ca 31 km) and are not used to rank between zones. No composite instability index is provided.
\item
  A Route is a single D8 line, following a fixed corridor (like a screen connection, as opposed to a runout, velocity or inundation model {[}45{]}). The published length of the Route to Tapovan was 2.5 km shorter than that actually constructed (Section 5.4).
\end{itemize}

\section{Conclusions and Future Work}\label{conclusions-and-future-work}

For the 2021 Chamoli event a single D8 route from the documented source was identified to pass all four documented checkpoints in sequence and had a corridor of approximately 250 m that encircled all three damaged sites; however the public asset inventory listed only Raini bridge as being affected by the event. A U-Net performed better than the spectral rules for surface mapping with a macro F1-score of 0.648 versus 0.519. For the class rock the spectral rules performed better whereas for the class visible ground the spectral rules were able to detect change from snow to ground that the U-Net could not. Under the declared volume rule the pipeline captured only 0.52 times the declared volume, and the energy estimate for only one of the zones (identified after the fact) fell within the range of numbers given in Section 5.1 for that zone. The northeast-facing source was also missed by the Winter Sentinel-2 change detection analysis, likely due to shadow suppression. A 2 m DEM difference (slope angle) is typically used post-event for detection of a release. Even with detection in place there is no warning given for an impending release. The pipeline now ranks the best zone within 1km of the source 14th of 552 (top 2.5\%), whereas a simple volume ranking of exposure before energy would place a zone near the source first or second. A single retrospective test case has been implemented using a set of tuned parameters to test on the same data. This work identifies two points of failure in the current pipeline, namely source detection in shadowed winter imagery and the ranking rule. However, this does not necessarily mean that the current pipeline is working well in general.

Next steps, in the order we will take them:

\begin{enumerate}
\def\labelenumi{\arabic{enumi}.}
\item
  Implement alternative source detection not depending on sunlight, e.g. by using Sentinel-1 SAR backscatter and change in coherence {[}7, 46{]} or even by using multi-date DEMs if available {[}8{]}.
\item
  Can a source be flagged before it fails using only pre-event data (e.g. multi-year DEM data or information on past changes of a glacier) so that the pipeline is used for early warning and not only for post-event analysis?
\item
  Modify the ranking rule from exposure first to a rule that combines energy with exposure and declare this new ranking rule before commencing the test.
\item
  Instead of a simple binary gate for illumination, use a graded confidence value and test on a different event.
\item
  Test the system on a second event and in non-event areas with change reference labels by a second analyst to measure false-alarm rates and ranking.
\item
  There are not enough labelled rock, soil and water examples in the feature map. The F1 values for the rock and soil feature maps stay around 0.3 and the water feature map has no labels.
\item
  Compare the energy for this single event with an independent estimate for the event. Compare the single D8 flow route used for the calculations with multiple-flow-direction routing as well as with a runout model {[}45{]} and state the distance errors for the individual checkpoints for which published distances for the Rishiganga and Tapovan Plants
\item
  Expose hydropower and critical infrastructure with an authoritative register of terms and definitions.
\end{enumerate}

\section{Data and Code Availability}\label{data-and-code-availability}

All code, configuration, labels, evaluation results, the follow-up protocol and the Kaggle notebooks can be found at https://github.com/Snigdho2005/mountain-hazard-screening. Sentinel-2 L2A data {[}5, 20{]} and Copernicus GLO-30 data {[}21{]} are both provided as open data by ESA/Copernicus. The 2 m DEMs are provided by Shean et al. {[}8{]} (CC BY-NC 4.0). ERA5 data {[}22{]} was downloaded via the Open-Meteo archive API {[}23{]}. The © OpenStreetMap data {[}24{]} used in this work is provided under the Open Database License (ODbL).

\section{Author Contributions and Use of AI Tools}\label{author-contributions-and-use-of-ai-tools}

This work was conducted by the author who conceived the work, designed the evaluation, created all reference labels, ran all experiments and analyzed resulting data. Generative AI was used to help write the work, to correct code, to revise the work, and to find and format references. The author of this paper reviewed and edited all parts written by AI, checked all references against the original sources, and is fully responsible for the contents of this paper: text, figures, data, references etc.

\section{}\label{section}

\section{References}\label{references}

{[}1{]} O. Hungr, S. Leroueil, and L. Picarelli. The Varnes classification of landslide types, an update. Landslides, 11(2):167-194, 2014. doi:10.1007/s10346-013-0436-y.

{[}2{]} A. Kääb, M. Jacquemart, A. Gilbert, S. Leinss, L. Girod, C. Huggel, D. Falaschi, F. Ugalde, D. Petrakov, S. Chernomorets, M. Dokukin, F. Paul, S. Gascoin, E. Berthier, and J. S. Kargel. Sudden large-volume detachments of low-angle mountain glaciers: more frequent than thought? The Cryosphere, 15(4):1751-1785, 2021. doi:10.5194/tc-15-1751-2021.

{[}3{]} D. H. Shugar et al. A massive rock and ice avalanche caused the 2021 disaster at Chamoli, Indian Himalaya. Science, 373(6552):300-306, 2021. doi:10.1126/science.abh4455.

{[}4{]} A. B. Shrestha, J. Steiner, S. Nepal, S. B. Maharjan, M. Jackson, G. Rasul, and B. Bajracharya. Understanding the Chamoli flood: cause, process, impacts, and context of rapid infrastructure development. ICIMOD, Kathmandu, March 3, 2021. \href{https://www.icimod.org/article/understanding-the-chamoli-flood-cause-process-impacts-and-context-of-rapid-infrastructure-development/}{Report}.

{[}5{]} M. Drusch et al. Sentinel-2: ESA\textquotesingle s optical high-resolution mission for GMES operational services. Remote Sensing of Environment, 120:25-36, 2012. doi:10.1016/j.rse.2011.11.026.

{[}6{]} O. Ghorbanzadeh, Y. Xu, P. Ghamisi, M. Kopp, and D. Kreil. Landslide4Sense: reference benchmark data and deep learning models for landslide detection. IEEE Transactions on Geoscience and Remote Sensing, 60:1-17, 2022. doi:10.1109/TGRS.2022.3215209.

{[}7{]} L. Nava, A. Mondini, K. Bhuyan, C. Fang, O. Monserrat, A. Novellino, and F. Catani. Sentinel-1 SAR-based globally distributed co-seismic landslide detection by deep neural networks. Geoscientific Model Development, 19(1):167-185, 2026. doi:10.5194/gmd-19-167-2026.

{[}8{]} D. Shean, S. Bhushan, E. Berthier, C. Deschamps-Berger, S. Gascoin, and F. Knuth. Chamoli disaster post-event 2-m DEM composite (February 10 and 11, 2021) and difference map. Zenodo, 2021. \href{https://zenodo.org/record/4558692}{doi:10.5281/zenodo.4558692}; and S. Bhushan and D. Shean. Chamoli disaster pre-event 2-m DEM composite: September 2015. Zenodo, 2021. \href{https://zenodo.org/record/4554647}{doi:10.5281/zenodo.4554647}.

{[}9{]} P. Vangla, D. Jindal, A. Sajwan, R. Devrani, J. D. Frost, and G. V. Ramana. Site visit and drone based reconnaissance study of the severely affected infrastructure by February 7th, 2021, Chamoli rock-ice avalanche disaster, Uttarakhand, India. GEER Association Report GEER-074, 2022. doi:10.18118/G6V95B. \href{https://geerassociation.org/components/com_geer_reports/geerfiles/Chamoli_Report.pdf}{Report}.

{[}10{]} O. Ronneberger, P. Fischer, and T. Brox. U-Net: convolutional networks for biomedical image segmentation. In MICCAI 2015, LNCS 9351, pages 234-241, 2015.

{[}11{]} M. Galatola, E. Arnaudo, L. Barco, C. Rossi, and F. Dominici. Land cover segmentation with sparse annotations from Sentinel-2 imagery. In IGARSS 2023, pages 6952-6955, 2023. doi:10.1109/IGARSS52108.2023.10281933.

{[}12{]} K. Jindgar and G. W. Lindsay. Improving satellite imagery segmentation using multiple Sentinel-2 revisits. arXiv preprint arXiv:2409.17363, 2024. doi:10.48550/arXiv.2409.17363.

{[}13{]} D. Lu, P. Mausel, E. Brondízio, and E. Moran. Change detection techniques. International Journal of Remote Sensing, 25(12):2365-2401, 2004. doi:10.1080/0143116031000139863.

{[}14{]} R. D. Johnson and E. S. Kasischke. Change vector analysis: a technique for the multispectral monitoring of land cover and condition. International Journal of Remote Sensing, 19(3):411-426, 1998. doi:10.1080/014311698216062.

{[}15{]} J. F. O\textquotesingle Callaghan and D. M. Mark. The extraction of drainage networks from digital elevation data. Computer Vision, Graphics, and Image Processing, 28(3):323-344, 1984.

{[}16{]} P. Horton, M. Jaboyedoff, B. Rudaz, and M. Zimmermann. Flow-R, a model for susceptibility mapping of debris flows and other gravitational hazards at a regional scale. Natural Hazards and Earth System Sciences, 13(4):869-885, 2013. doi:10.5194/nhess-13-869-2013.

{[}17{]} J. Höhle and M. Höhle. Accuracy assessment of digital elevation models by means of robust statistical methods. ISPRS Journal of Photogrammetry and Remote Sensing, 64(4):398-406, 2009.

{[}18{]} P. Olofsson, G. M. Foody, M. Herold, S. V. Stehman, C. E. Woodcock, and M. A. Wulder. Good practices for estimating area and assessing accuracy of land change. Remote Sensing of Environment, 148:42-57, 2014.

{[}19{]} N. Khadka, X. Chen, Y. Nie, S. Thakuri, G. Zheng, and G. Zhang. Evaluation of glacial lake outburst flood susceptibility using multi-criteria assessment framework in Mahalangur Himalaya. Frontiers in Earth Science, 8:601288, 2021. doi:10.3389/feart.2020.601288.

{[}20{]} M. Main-Knorn, B. Pflug, J. Louis, V. Debaecker, U. Müller-Wilm, and F. Gascon. Sen2Cor for Sentinel-2. In Proc. SPIE 10427, Image and Signal Processing for Remote Sensing XXIII, 1042704, 2017. doi:10.1117/12.2278218.

{[}21{]} European Space Agency and Airbus. Copernicus DEM. 2022. doi:10.5270/ESA-c5d3d65.

{[}22{]} H. Hersbach et al. The ERA5 global reanalysis. Quarterly Journal of the Royal Meteorological Society, 146(730):1999-2049, 2020.

{[}23{]} P. Zippenfenig. Open-Meteo.com Weather API. Zenodo, 2023. doi:10.5281/zenodo.7970649.

{[}24{]} M. Haklay and P. Weber. OpenStreetMap: user-generated street maps. IEEE Pervasive Computing, 7(4):12-18, 2008. doi:10.1109/MPRV.2008.80.

{[}25{]} QGIS Development Team. QGIS Geographic Information System. Open Source Geospatial Foundation Project. https://qgis.org (accessed 2026).

{[}26{]} D. K. Hall, G. A. Riggs, and V. V. Salomonson. Development of methods for mapping global snow cover using Moderate Resolution Imaging Spectroradiometer data. Remote Sensing of Environment, 54(2):127-140, 1995.

{[}27{]} J. W. Rouse, R. H. Haas, J. A. Schell, and D. W. Deering. Monitoring vegetation systems in the Great Plains with ERTS. In Proceedings of the Third ERTS Symposium, NASA SP-351, volume 1, pages 309-317, 1974.

{[}28{]} S. K. McFeeters. The use of the Normalized Difference Water Index (NDWI) in the delineation of open water features. International Journal of Remote Sensing, 17(7):1425-1432, 1996.

{[}29{]} M. Guizar-Sicairos, S. T. Thurman, and J. R. Fienup. Efficient subpixel image registration algorithms. Optics Letters, 33(2):156-158, 2008.

{[}30{]} L. W. Zevenbergen and C. R. Thorne. Quantitative analysis of land surface topography. Earth Surface Processes and Landforms, 12(1):47-56, 1987.

{[}31{]} I. Loshchilov and F. Hutter. Decoupled weight decay regularization. In International Conference on Learning Representations (ICLR), 2019. arXiv:1711.05101.

{[}32{]} I. Loshchilov and F. Hutter. SGDR: stochastic gradient descent with warm restarts. In International Conference on Learning Representations (ICLR), 2017. arXiv:1608.03983.

{[}33{]} M. Iqbal. An Introduction to Solar Radiation. Academic Press, New York, 1983.

{[}34{]} A. Saltelli, M. Ratto, T. Andres, F. Campolongo, J. Cariboni, D. Gatelli, M. Saisana, and S. Tarantola. Global Sensitivity Analysis: The Primer. Wiley, Chichester, 2008.

{[}35{]} C. Spearman. The proof and measurement of association between two things. American Journal of Psychology, 15(1):72-101, 1904.

{[}36{]} C. Rolstad, T. Haug, and B. Denby. Spatially integrated geodetic glacier mass balance and its uncertainty based on geostatistical analysis: application to the western Svartisen ice cap, Norway. Journal of Glaciology, 55(192):666-680, 2009. doi:10.3189/002214309789470950.

{[}37{]} K. M. Cuffey and W. S. B. Paterson. The Physics of Glaciers. Butterworth-Heinemann, Oxford, 4th edition, 2010.

{[}38{]} L. Vincent. Morphological grayscale reconstruction in image analysis: applications and efficient algorithms. IEEE Transactions on Image Processing, 2(2):176-201, 1993.

{[}39{]} M. Sokolova and G. Lapalme. A systematic analysis of performance measures for classification tasks. Information Processing \& Management, 45(4):427-437, 2009.

{[}40{]} C. J. Clopper and E. S. Pearson. The use of confidence or fiducial limits illustrated in the case of the binomial. Biometrika, 26(4):404-413, 1934.

{[}41{]} C. R. Harris et al. Array programming with NumPy. Nature, 585:357-362, 2020.

{[}42{]} P. Virtanen et al. SciPy 1.0: fundamental algorithms for scientific computing in Python. Nature Methods, 17:261-272, 2020.

{[}43{]} S. van der Walt, J. L. Schönberger, J. Nunez-Iglesias, F. Boulogne, J. D. Warner, N. Yager, E. Gouillart, T. Yu, and the scikit-image contributors. scikit-image: image processing in Python. PeerJ, 2:e453, 2014.

{[}44{]} A. Paszke et al. PyTorch: an imperative style, high-performance deep learning library. In Advances in Neural Information Processing Systems 32, pages 8024-8035, 2019.

{[}45{]} M. Mergili, J.-T. Fischer, J. Krenn, and S. P. Pudasaini. r.avaflow v1, an advanced open-source computational framework for the propagation and interaction of two-phase mass flows. Geoscientific Model Development, 10(2):553-569, 2017. doi:10.5194/gmd-10-553-2017.

{[}46{]} R. Torres et al. GMES Sentinel-1 mission. Remote Sensing of Environment, 120:9-24, 2012.

{[}47{]} P. J. Rousseeuw and C. Croux. Alternatives to the median absolute deviation. Journal of the American Statistical Association, 88(424):1273-1283, 1993. doi:10.1080/01621459.1993.10476408.

{[}48{]} C. Leys, C. Ley, O. Klein, P. Bernard, and L. Licata. Detecting outliers: do not use standard deviation around the mean, use absolute deviation around the median. Journal of Experimental Social Psychology, 49(4):764-766, 2013. doi:10.1016/j.jesp.2013.03.013.

\end{document}